%% file: continuum.tex
\documentclass{article}

\usepackage{arxiv}

\usepackage[utf8]{inputenc} 
\usepackage[T1]{fontenc}    
\usepackage{hyperref}       
\usepackage{url}            
\usepackage{booktabs}       
\usepackage{amsfonts}       
\usepackage{nicefrac}       
\usepackage{microtype}      
\usepackage[pdftex]{graphicx}
\usepackage{amssymb,amsfonts,amsmath,amscd}
\usepackage[printonlyused,withpage]{acronym}
\usepackage{natbib}
\usepackage{mathrsfs}
\usepackage{xcolor}
\usepackage{soul}

\graphicspath{{figs/}}
\input{notation_header}
\acrodef{BA}{Bundle Adjustment}
\acrodef{DNN}{Deep Neural Network}
\acrodef{EM}{Expectation Maximization}
\acrodef{GEM}{Generalized Expectation Maximization}
\acrodef{HMM}{Hidden Markov Model}
\acrodef{LDS}{Linear Dynamical System}
\acrodef{LQG}{Linear Quadratic Gaussian}
\acrodef{LQR}{Linear Quadratic Regulator}
\acrodef{LTI}{Linear Time-Invariant}
\acrodef{RTS}{Rauch-Tung-Striebel}
\acrodef{SGD}{Stochastic Gradient Descent}
\acrodef{SLAM}{Simultaneous Localization and Mapping}
\acrodef{RKHS}{Reproducing Kernel Hilbert Space}
\acrodef{SMW}{Sherman-Morrison-Woodbury}

\acrodef{GVI}{Gaussian Variational Inference}
\acrodef{ESGVI}{Exactly Sparse Gaussian Variational Inference}
\acrodef{MAP}{Maximum A Posteriori}
\acrodef{ML}{Maximum Likelihood}
\acrodef{KL}{Kullback-Leibler}
\acrodef{PDF}{Probability Density Function}
\acrodef{NEES}{Normalized Estimation Squared Error}
\acrodef{KF}{Kalman Filter}
\acrodef{VKF}{Variational Kalman Filter}
\acrodef{ISPKF}{Iterated Sigmapoint Kalman Filter}
\acrodef{ESGVI-GN}{ESGVI Gauss-Newton}
\acrodef{ELBO}{Evidence Lower Bound}
\acrodef{NGD}{Natural Gradient Descent}
\acrodef{FIM}{Fisher Information Matrix}
\acrodef{RANSAC}{Random Sample And Consensus}
\acrodef{IRLS}{Iteratively Reweighted Least-Squares}
\acrodef{BRD}{Black-Rangarajan Duality}
\acrodef{GNC}{Graduated Non-Convexity}
\acrodef{GP}{Gaussian process}

\hypersetup{%
    pdftitle={},
    pdfauthor={},
    pdfkeywords={},
    pdfsubject={},
    pdfstartview=FitH,%
    bookmarks=true,%
    bookmarksopen=true,%
    breaklinks=true,%
    colorlinks=true,%
    linkcolor=black,
    anchorcolor=black,%
    citecolor=black,
    filecolor=black,%
    menucolor=black,
    pagecolor=black,%
    urlcolor=black
}%

\title{Continuum Robot State Estimation Using \\ Gaussian Process Regression on $SE(3)$}

\author{
	\normalfont Sven Lilge$^{1,2}$
	\and
 \normalfont Timothy D. Barfoot$^{1,3}$
 \and
 \normalfont Jessica Burgner-Kahrs$^{1,2}$ 
 \thanks{\hspace*{-1.8em}$^1$Robotics Institute, University of Toronto, Toronto, ON, Canada \newline
 	$^2$Continuum Robotics Laboratory, Department of Mathematical \&
 	Computational Sciences, University of Toronto, Mississauga, ON,
 	Canada \newline
 	$^3$Autonomous Space Robotics Laboratory, Institute for Aerospace Studies, University of Toronto, Toronto, ON, Canada \newline
 	Corresponding author: Sven Lilge, e-mail: \texttt{slilge@cs.toronto.edu}}
}

\date{}

\renewcommand\footnotemark{}

\begin{document}

\maketitle
\title{Continuum Robot State Estimation Using  Gaussian Process Regression on $SE(3)$}

\begin{abstract}
Continuum robots have the potential to enable new applications in medicine, inspection, and countless other areas due to their unique shape, compliance, and size.  Excellent progess has been made in the mechanical design and dynamic modelling of continuum robots, to the point that there are some canonical designs, although new concepts continue to be explored. In this paper, we turn to the problem of state estimation for continuum robots that can been modelled with the common Cosserat rod model.  Sensing for continuum robots might comprise external camera observations, embedded tracking coils or strain gauges.  We repurpose a \ac{GP} regression approach to state estimation, initially developed for continuous-time trajectory estimation in $SE(3)$.  In our case, the continuous variable is not time but arclength and we show how to estimate the continuous shape (and strain) of the robot (along with associated uncertainties) given discrete, noisy measurements of both pose and strain along the length.  We demonstrate our approach quantitatively through simulations as well as through experiments.
Our evaluations show that accurate and continuous estimates of a continuum robot's shape can be achieved, resulting in average end-effector errors between the estimated and ground truth shape as low as 3.5mm and 0.016$^\circ$ in simulation or 3.3mm and 0.035$^\circ$ for unloaded configurations and 6.2mm and 0.041$^\circ$ for loaded ones during experiments, when using discrete pose measurements.

\end{abstract}

\keywords{continuum robot \and matrix Lie groups \and state estimation \and Gaussian process regression}

\section{Introduction}
Continuum robots are actuatable, compliant structures, whose constitutive material forms curves with continuous tangent vectors. 
The motion capabilities of continuum robots can be expressed by a set of motion primitives (extension/contraction, bending, shear, and twisting) that are realized in concatenated segments composing the robot.
The compliance inherent in the continuum structure allows the robot to adapt to contact with its environment and to compensate for the limitations in its actuatable degrees of freedom.
At small diameters and high lengths, continuum robots show great potential for manipulation in constrained environments such as medical \citep{Burgner-Kahrs2015a} and industrial applications, e.g.\ in-situ inspection, maintenance, and repair \citep{Dong2019,Wang2021}.

We have seen significant advancements on how to effectively design and actuate small-scale continuum robots over the past two decades \citep{Webster2010,Walker2013a}.
In order to efficiently plan and control motions of these manipulators, usually accurate models are required.
While the modelling accuracy achieved by the most sophisticated approaches for continuum robots is promising \citep{Rucker2010,rucker2011statics}, non-negligible discrepancies between model predictions and robot prototypes persist, largely due to unmodelled effects.
As these inaccuracies make accurate control and motion planning challenging, sensors are used to infer information about the robot's actual state.
In continuum robotics research, we predominantly use discrete position or pose information, usually obtained from external cameras \citep{Hannan2005} or embedded electromagnetic tracking coils \citep{Mahoney2016}, as well as discrete strain information, obtained from embedded fiber Bragg grating sensors \citep{Modes2021}.
An overview of different shape sensing techniques for continuum robots, focusing on their possible application in minimally invasive surgery, is provided by \cite{Shi2017}. \cite{Mahoney2018a} emphasize that sensing is a crucial component in the deployment of continuum robots, being fundamentally coupled to their design, motion planning, and control.
Therefore, special attention should be given to accurately estimating the state of continuum robots based on sensor readings.

\begin{table}[b!]\renewcommand*{\arraystretch}{1.25}
	\centering
	\caption{Overview of related work on state estimation for continuum robots}
	\small
	\begin{tabular}{p{3.5cm} p{4cm} p{4cm} p{3cm}}
		\hline Reference  & Robot Representation \newline and Model & State Estimation Approach & Estimated State \newline Variables \\
		\hline\hline
		\multicolumn{4}{c}{\textbf{Shape fitting and optimization based methods}} \\
		\cite{Roesthuis2013} & Piecewise constant curvature & Defining constant curvature arcs from strain measurements & Pose along arclength \\
		\cite{Kim2014} & Polynomial curvature functions & Curvature regression & Pose along arclength \\
		\cite{Song2015} & Shape as B\'{e}zier curve & Shape curve fitting & Pose along arclength \\
		\cite{Rone2013}  & Piecewise constant curvature with virtual power mechanics & Cable displacement fitting & Pose along arclength and actuation forces\\
		\cite{Venkiteswaran2019}  & Pseudo rigid body with Newton-Euler mechanics & Shape fitting & Pose along arclength and tip force \\
		\hline
		\multicolumn{4}{c}{\textbf{Statistical estimation methods}} \\
		\cite{Brij2010}& Constant curvature kinematics  & Unscented particle filter& Tip pose\\
		\cite{Borgstadt2015} & Constant curvature kinematics & Dual particle filter & Tip position \\
		\cite{Chen2019} & Constant curvature kinematics & Unscented Kalman filter & Constant curvature arc parameter\\
		\cite{Ataka2016} & Constant curvature kinematics & Extended Kalman filter & Actuation variables \\
		\cite{Loo2019} & Constant curvature kinematics & Extended Kalman filter & Constant curvature arc parameter \\
		\cite{Lobaton2013} & Learned shape basis functions & Kalman filter & Shape basis function \newline coefficients\\
		\cite{Mahoney2016} and \cite{Anderson2017}  & Kirchhoff rod mechanics & Kalman-Bucy filter & Pose and internal stress along arclength\\
		Our method & Cosserat rod mechanics & Gaussian process regression & Pose and strain along \newline  arclength \\
		\hline
	\end{tabular}
	\label{tab:sota_overview}
\end{table}

One commonly used approach to infer information about a continuum robot's shape is to use optimization methods to fit a given shape representation to discrete sensor observations.
This can either mean fitting the parameters of a mechanics-based kinematic model to align the modelled shape with external sensor information, or directly fitting shape representation functions, such as polynomials or other continuous curves to the discrete sensor readings.
Examples include shape estimation approaches utilizing models based on the principle of virtual power \citep{Rone2013} and pseudo rigid-body assumptions \citep{Venkiteswaran2019}, as well as finding Bezier curves \citep{Song2015}, polynomials \citep{Kim2014} or constant curvatures \citep{Roesthuis2013} that match the measurements.
While these approaches make it possible to reason about the shape of the continuum robot,
sensor noise as well as modelling uncertainties are not explicitly taken into account.
Thus, more sophisticated stochastic state estimation approaches based on statistical methods are desirable.

The majority of approaches that apply statistical methods to estimate the robot's end-effector pose or shape are based on simplified kinematic models that assume bending in constant curvatures.
Examples include, shape estimation of catheters \citep{Brij2010,Borgstadt2015}, tendon-driven continuum robots \citep{Chen2019,Ataka2016} and soft robots \citep{Loo2019}.
In another approach, \cite{Lobaton2013} incorporate a more general shape representation, consisting of a combination of spatial basis functions, with a stochastic state estimation approach.
All of these approaches provide filtered and robust estimates of the robot's shape given noisy sensor data.
However, due to the simplified geometrically motivated assumptions, their estimated state is restricted to only include the position and orientation of the robot's shape.
Thus, it does not include internal strains, which could be used to gain insights into the robot's internal forces and moments as well as to estimate external loads.
A more complete state representation of continuum robots to be used in stochastic state estimation approaches is proposed by \cite{Mahoney2016}.
In their work, the state of a continuum robot is based on a differential formulation, allowing to represent more sophisticated mechanics-based kinematic models.
Based on this state representation, an estimation approach utilizing a Kalman-Bucy filter in combination with a Rauch-Tung-Striebel smoother is proposed.
The approach is applied to a concentric tube continuum robot and a parallel continuum robot setup in \citep{Anderson2017}, using a Kirchhoff rod modelling approach in both cases.

Following a similar approach, we are proposing a general state estimation method for continuum robots that can be modelled using Cosserat rod theory.
The proposed state estimation approach is based on a sparse Gaussian process regression, a method previously applied to continuous-time trajectory estimation in $SE(3)$ \citep{barfoot_rss14, anderson_iros15, barfoot_ser17}.
By exploiting the similarities of the Cosserat rod model to the dynamics of a rigid body \citep{deleuterio85}, the method is repurposed for continuous state estimation along a continuum robot's arclength, $s$.
The estimated state includes both the pose and the internal strain variables along the robot's structure and values can easily be interpolated between discrete estimation nodes to obtain a continuous representation including uncertainty envelopes.
The framework follows a general Cosserat rod formulation, allowing application to any continuum robot and requiring no robot-dependent kinematic model or particular knowledge about the robot's properties, actuation inputs, or external forces.
For evaluation, the proposed state estimation framework is applied to a tendon driven continuum robot, including both simulations and experiments using a multisegment manipulator, as an example.

With respect to the related work on continuum robot state estimation (overview in Table~\ref{tab:sota_overview}) or method is closest to the approaches of \cite{Mahoney2016} and \cite{Anderson2017}, as it employs a similar mechanics-based Cosserat rod model to estimate both the pose and strain variables along the robot's arclength.
However, there are some important differences that set our work apart.
First, we are using Gaussian process regression instead of using a filtering approach.
This method has the advantage that it can compute an estimate for all robot states at the same time, taking into account model uncertainties and measurements, while filtering approaches usually require to iteratively integrate along the robot's arclength.
Second, we are deriving and employing a simple and general prior model, which leads to a sparse structure for the Gaussian process regression, allowing efficient computations.

Thus, the main contribution and novelty of our work are its versatility, allowing it to be applied to any Cosserat rod type continuum robot, and its potential for computational efficiency.

\section{Theory}

Our approach to state estimation for continuum robots is to fuse a prior distribution over robot states (i.e., shape and strain from a physical model) with sensed quantities (i.e., from camera or strain gauges) to generate a posterior distribution over robot states.  In other words, we will take a probabilistic approach in order to estimate not only the most likely robot state but also its uncertainty.  Philosophically, we do not require the most accurate prior model since the measurements will be tremendously helpful in our task.  We therefore develop and employ a simplified Cosserat rod model that essentially allows us to `interpolate' between discrete measurements.  This section first develops this model then shows how to use it for continuous state estimation via sparse \acf{GP} regression. We will follow the notation of \citet{barfoot_ser17} to describe quantities in $SE(3)$. Table~\ref{tab:nomenclature} summarizes the nomenclature used in this paper.
\subsection{Cosserat Rod Model}

\begin{figure}[t]
\centering
\includegraphics[width=\textwidth]{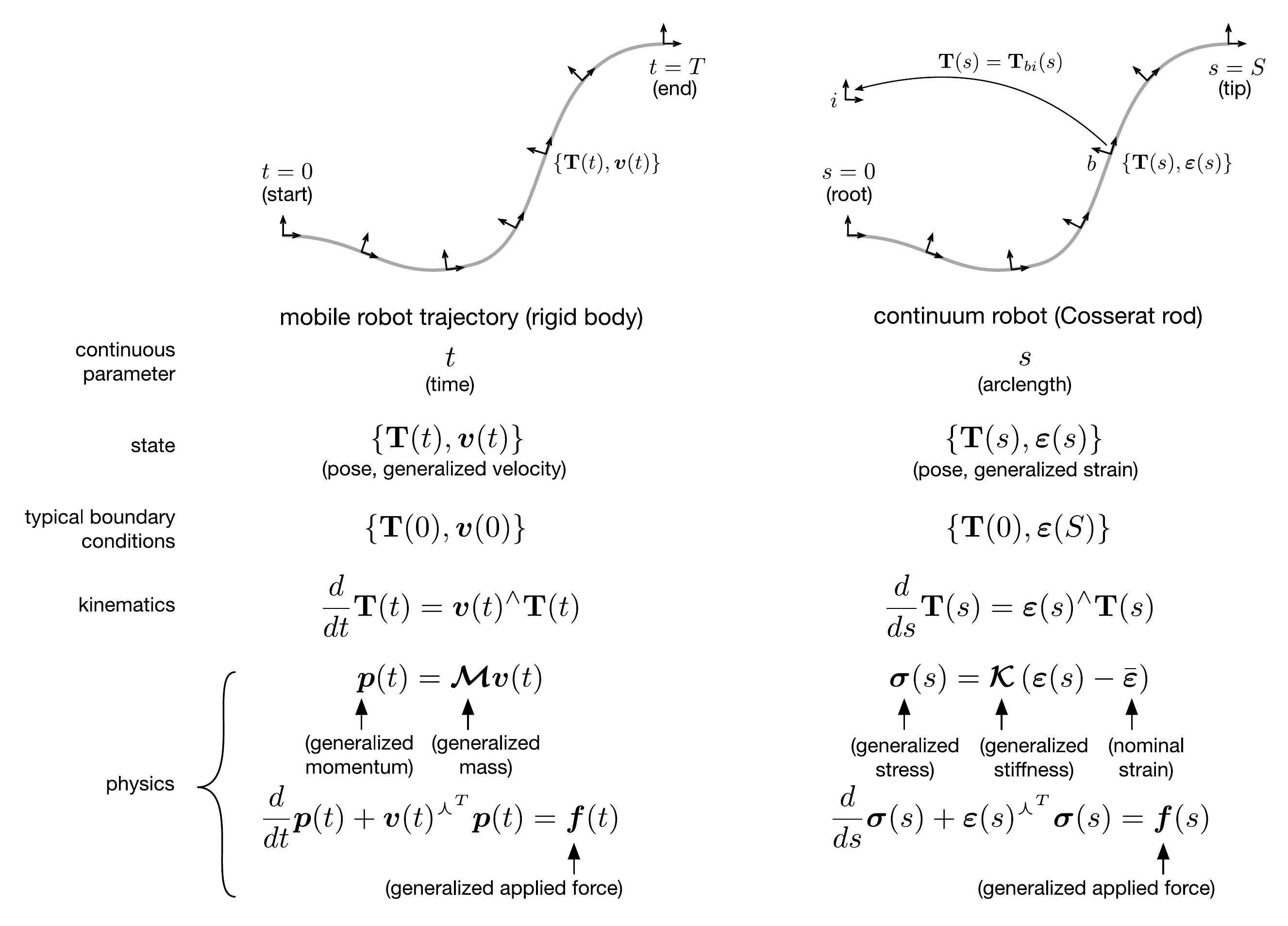}
\caption{Comparison of the equations of motion (parameterized by time) of a rigid body \citep{deleuterio85} to those of the quasi-static Cosserat rod model (parameterized by arclength) \citep{pai02}. Quantities are expressed in the body frame and $\mbf{T} = \mbf{T}_{bi}$, i.e., it transforms quantities from a static frame, $i$, to the body frame, $b$. }
\label{fig:model}
\end{figure}

We seek a reasonable model of the physics of a continuum robot to serve as our prior for state estimation.  The well-known Cosserat rod model, commonly employed in continuum robotics \citep{rucker2011statics,rao2021model}, serves the purpose.  Simplifications of the Cosserat rod are commonly employed in the graphics community \citep{pai02}, and we will leverage a similar version here.
Figure~\ref{fig:model} notes the similarity of the Cosserat rod model to the dynamics of a rigid body, when expressed in $SE(3)$ \citep{deleuterio85}, an analogy whose idea was first discussed by \cite{kirchhoff1859ueber}.

\begin{table}[ht!]\renewcommand*{\arraystretch}{1.25}
	\centering
	\caption{Nomenclature}
	\small
	\begin{tabular}{l p{4.75cm} l p{4.75cm} }
		\hline \multicolumn{2}{c}{Cosserat Rod Model} & \multicolumn{2}{c}{Gaussian Process Regression}  \\
		\hline\hline$\mbf{x}(s)$ & State over arclength $s$ & $K$ & Number of discrete arclength positions $s_k$ \\
		$i$ & Static (inertial) frame & $\mbf{x}(s_k) = \mbf{x}_k$ & State at discrete arclength $s_k$ \\
		$b$ & Body frame & $\pri{\mbf{x}}(s)$, $\pri{\mbf{P}}(s,s^\prime)$ & Prior mean and covariance functions \\
		$\mbf{T}_{ib}(s)$, $\mbf{T}_{bi}(s) = \mbf{T}(s)$ & Transformation from body to static frame and vice versa; lack of subscript indicates the latter & $\mbf{w}(s)$ & White-noise Gaussian process \\
		$\mbs{\varepsilon}_i(s)$, $\mbs{\varepsilon}_b(s) = \mbs{\varepsilon}(s)$ & Generalized strain in static and body frame; lack of subscript indicates the latter & 	$\mbf{Q}_c(s)$ & Covariance function of $\mbf{w}(s)$ \\
		$\mbf{C}(s)$ & Rotation of the pose & 	$\mbs{\xi}_k(s)$ & Local pose variable in the Lie algebra \\
		$\mbf{r}(s)$ & Translation of the pose & 	$\mbs{\psi}_k(s)$ & Derivative of local pose \\
		$\Tbig(s)$ & Adjoint of the pose & $\mbs{\gamma}_k(s)$ & Markovian state\\ $\bar{\mbs{\varepsilon}}(s)$ & Nominal strain of stress-free rod & $\mbf{e}_{p,k}$, $\mbf{e}_{m,k}$ & Error terms for prior and measurements \\
		$\mbs{\nu}(s)$ & Translational generalized strain & $J_{p,k}$, $J_{m,k}$ & Cost terms for prior and measurements \\
		$\mbs{\omega}(s)$ & Rotational generalized strain & $\Jbig(s)$ & Left Jacobian of $SE(3)$\\
		$\mbs{\sigma}(s)$& Generalized stress & 	$\tilde{\mbf{T}}_k$, $\tilde{\mbs{\varepsilon}}_k$ & Sample drawn from Gaussian distribution related to measurements\\
		$\mbs{\mathcal{K}}(s)$ & Generalized stiffness& 	${\mbf{R}}_k$ & Covariance of Gaussian distribution related to measurements\\
		$\mbs{f}(s)$ & External forces & 	$\mbf{E}_{p,k}$, $\mbf{E}_{m,k}$ & Error Jacobians for prior and measurements\\
		\hline
	\end{tabular}
	\label{tab:nomenclature}
\end{table}

We consider the {\em state} of the continuum robot to be
\begin{equation}
\mbf{x}(s) = \{ \mbf{T}(s), \mbs{\varepsilon}(s) \},
\end{equation}
where $\mbf{T}(s) \in SE(3)$ provides the six-degree-of-freedom {\em pose}, $\mbs{\varepsilon}(s) \in \mathbb{R}^6$ describes the six-degree-of-freedom {\em generalized strain}, and both are parameterized continuously as a function of arclength coordinate, $s$.  Although we will avoid frame subscripts to keep the notation clean, the lack of subscripts will imply $\mbf{T}(s) = \mbf{T}_{bi}(s)$ and $\mbs{\varepsilon}(s) = \mbs{\varepsilon}_{b}(s)$, with $i$ a static (inertial) frame and $b$ the body frame.  This is the state we will eventually want to estimate, although ultimately we will have to be satisfied with querying it at a finite number of values for the arclength.  

At times we will also require the {\em adjoint} of the pose, $\Tbig(s) \in \mbox{Ad}(SE(3))$,
\begin{equation}
\mbf{T}(s) = \bbm \mbf{C}(s) & \mbf{r}(s) \\ \mbf{0}^T & 1 \ebm, \qquad \Tbig(s) = \mbox{Ad}(\Tsmall(s)) = \bbm \mbf{C}(s) & \mbf{r}(s)^\wdg \mbf{C}(s) \\ \mbf{0} & \mbf{C}(s) \ebm,
\end{equation}
where $\mbf{C}(s) \in SO(3)$ is rotation, $\mbf{r}(s) \in \mathbb{R}^3$ is translation, and the usual skew-symmetric operator  $\wdg$  is given by
\begin{equation}
\mbf{r}^\wdg = \bbm r_1 \\ r_2 \\ r_3 \ebm^\wdg = \bbm 0 & -r_3 & r_2 \\ r_3 & 0 & -r_1 \\ -r_2 & r_1 & 0 \ebm,
\end{equation}
which can be used to implement the cross product.

We can further decompose the generalized strain into its translational, $\mbs{\nu}(s)$, and rotational, $\mbs{\om}(s)$, components according to 
\begin{equation}
\mbs{\varepsilon}(s) = \bbm \mbs{\nu}(s) \\ \mbs{\om}(s) \ebm. 
\end{equation}
It is worth noting that the analogue of the generalized strain in the continuous-time case is generalized velocity; here we must think of it as how quickly each component of the pose is changing as we proceed along the length of the rod.  

We can relate the pose to generalized strain through the {\em spatial kinematics} equation
\begin{equation}\label{eq:kinematics}
\frac{d}{ds}\mbf{T}(s) = \mbs{\varepsilon}(s)^\wdg \mbf{T}(s) \qquad \mbox{or}  \qquad \frac{d}{ds}\Tbig(s) = \mbs{\varepsilon}(s)^\Wdg \Tbig(s),
\end{equation}
where we overload the $\wdg$ operator and define the $\Wdg$ operator as
\begin{equation}
\mbs{\varepsilon}(s)^\wdg = \bbm \mbs{\om}(s)^\wdg & \mbs{\nu}(s) \\ \mbf{0}^T & 0 \ebm, \qquad \mbs{\varepsilon}(s)^\Wdg = \bbm \mbs{\om}(s)^\wdg & \mbs{\nu}(s)^\wdg  \\ \mbf{0} & \mbs{\om}(s)^\wdg \ebm.
\end{equation}
Intuitively, if the strain is known, we can integrate from the root of the robot to the tip to determine the pose along the length.

The {\em physics} equation (i.e., involving forces) requires a bit more explanation.  In an inertial (i.e., stationary) frame, $i$, the {\em generalized stress} can be obtained using the constitutive law
\begin{equation}
\mbs{\sigma}_i(s) = \mbs{\mathcal{K}}_i(s) \left( \mbs{\varepsilon}_i(s) - \bar{\mbs{\varepsilon}}_i(s) \right),
\end{equation}
where $\mbs{\mathcal{K}}_i(s)$ is the {\em generalized stiffness} and $\bar{\mbs{\varepsilon}}_i(s)$ is the {\em nominal generalized strain} (i.e., used to provide the nominal longitudinal length of the rod and as well as any pre-curvature).
In this context, $\mbs{\sigma}_i(s)$ is a six-dimensional vector expressing the robot's internal reaction forces and moments associated with bending, twisting shear and elongation. $\mbs{\mathcal{K}}_i(s)$ is a square stiffness matrix with respect to these deformations and  relates the strain variables to the reaction forces and moments.
We can relate rate of change of generalized stress to the applied force:
\begin{equation}
\frac{d}{ds} \mbs{\sigma}_i(s) = \mbs{f}_i(s).
\end{equation}
The trouble with writing this in the inertial frame is that the stiffness and nominal strain depend on the configuration of the robot, $\mbf{T}(s)$.  Skipping some of the details, we can transform the stress from the body (i.e., local) frame $b$ to the inertial frame $i$ as follows:
\begin{equation}
\mbs{\sigma}_i(s) = \Tbig_{ib}(s)^{-T} \mbs{\sigma}_b(s) = \Tbig_{ib}(s)^{-T} \mbs{\mathcal{K}}_b \left( \mbs{\varepsilon}_b(s) - \bar{\mbs{\varepsilon}}_b \right) ,
\end{equation}
where we now assume the stiffness and the nominal strain can be written as a constant with respect to arclength in the body frame, $b$ (this assumption could be relaxed but meets our needs for now).  We can then take the derivative (with respect to arclength) of both sides to reveal
\begin{equation}
\frac{d}{ds} \mbs{\sigma}_i(s) =  \left(  \frac{d}{ds}\Tbig_{ib}(s)^{-1} \right)^T \mbs{\sigma}_b(s) + \Tbig_{ib}(s)^{-T} \left( \frac{d}{ds} \mbs{\sigma}_b(s) \right),
\end{equation}
where we have applied the product rule of differentiation to the right side.  Inserting the kinematics~\eqref{eq:kinematics} for the pose derivative and manipulating we arrive at
\begin{equation}
\frac{d}{ds} \mbs{\sigma}_b(s) + \mbs{\varepsilon}_b(s)^{\Wdg^T} \mbs{\sigma}_b(s) = \mbs{f}_b(s). 
\end{equation}
Dropping all frame subscripts we have the following for the physics:
\beqn{dynamics}
\mbs{\sigma}(s) & = & \mbs{\mathcal{K}} \left( \mbs{\varepsilon}(s) - \bar{\mbs{\varepsilon}} \right), \\
\frac{d}{ds} \mbs{\sigma}(s)  + \underbrace{\mbs{\varepsilon}(s)^{\Wdg^T} \mbs{\sigma}(s)}_{\rm nonlinear} & = & \mbs{f}(s),
\eeqn
where we note the second term on the left in the bottom equation is nonlinear. 

Together, the kinematics~\eqref{eq:kinematics} and physics~\eqref{eq:dynamics} constitute our model of a continuum robot that we seek to use to regularize our state estimation problem.  The next section introduces the state estimation approach.

\subsection{Gaussian Process Regression for $SE(3)$}

In this section, we will repurpose the batch continuous-time state estimation framework first described by \citet{barfoot_rss14} and later made specific to $SE(3)$ by \citet{anderson_iros15,anderson_phd16}.  Intuitively, the \ac{GP} approach \citep{rasmussen06} allows us to perform estimation in the space of continuous functions, here the state of the robot given by $\mbf{x}(s) = \{ \mbf{T}(s), \mbs{\varepsilon}(s) \}$.  We actually have a fairly simple \ac{GP} problem since our input, $s$, is one dimensional.  Our kinematics/physics model will allow us to define a prior distribution over the possible states (i.e., shapes and strains) the robot can take, and this will be captured as a Gaussian process with mean and covariance functions.  Through Bayesian inference, discrete measurements of pose and strain along the robot length will then allow us to infer a posterior distribution, also a Gaussian process.  

\subsubsection{Prior}

Ideally, our \ac{GP} prior would take the form
\begin{equation}\label{eq:prior}
\mbf{x}(s) \sim \mathcal{GP}( \pri{\mbf{x}}(s), \pri{\mbf{P}}(s,s^\prime)), 
\end{equation}
where $\pri{\mbf{x}}(s)$ and $\pri{\mbf{P}}(s,s^\prime)$ are the prior mean and covariance functions, respectively.   The idea is to assign these prior mean and covariance functions using the kinematics/physics model from the previous section, corrupted by some process noise.  

Towards this end, the kinematics~\eqref{eq:kinematics} and physics~\eqref{eq:dynamics} from the previous section can be simplified as follows.  First, the physics can be written as
\begin{equation}
\frac{d}{ds} \left(  \mbs{\mathcal{K}} \left( \mbs{\varepsilon}(s) - \bar{\mbs{\varepsilon}} \right)  \right) = \mbs{f}(s)-\mbs{\varepsilon}(s)^{\Wdg^T} \mbs{\sigma}(s),
\end{equation}
or
\begin{equation}
\frac{d}{ds}  \mbs{\varepsilon}(s) = \underbrace{\mbs{\mathcal{K}}^{-1}(\mbs{f}(s)-\mbs{\varepsilon}(s)^{\Wdg^T} \mbs{\sigma}(s))}_{\rm unknown}. \label{eq:ds_strain}
\end{equation}
We take the fairly extreme position for now that the right hand side of~\eqref{eq:ds_strain} is unknown and as such we will replace it with a white-noise process, resulting in the following model:
\beqn{model}
\frac{d}{ds}\mbf{T}(s) & = & \mbs{\varepsilon}(s)^\wdg \mbf{T}(s), \qquad \mbf{T}(0) = \mbf{1},\\
\frac{d}{ds} \mbs{\varepsilon}(s) & = & \mbf{w}(s), \qquad \mbf{w}(s) \sim \mathcal{GP}( \mbf{0}, \mbf{Q}_c(s-s^\prime)), \qquad \mbs{\varepsilon}(S) = \bar{\mbs{\varepsilon}},
\eeqn
where $\mbf{w}(s)$ is a white-noise Gaussian process with zero mean function and covariance function $ \mbf{Q}_c(s-s^\prime)$; $\mbf{Q}_c$ is a stationary power-spectral density matrix (i.e., continuous version of covariance matrix).  For the boundary conditions, we constrain the root pose to the identity transform and the tip strain to the nominal strain.  If we actually know something about the applied forces, we could capture this knowledge in a nonzero mean function.  The covariance function essentially allows us to capture how smooth we expect the state of the robot to be.  We have incorporated the stiffness, $\mbs{\mathcal{K}}$, into $\mbf{Q}_c$ so that this quantity alone now serves as the knob by which we can tune the robot smoothness.

\begin{figure}[t]
\centering
\includegraphics[width=0.55\textwidth]{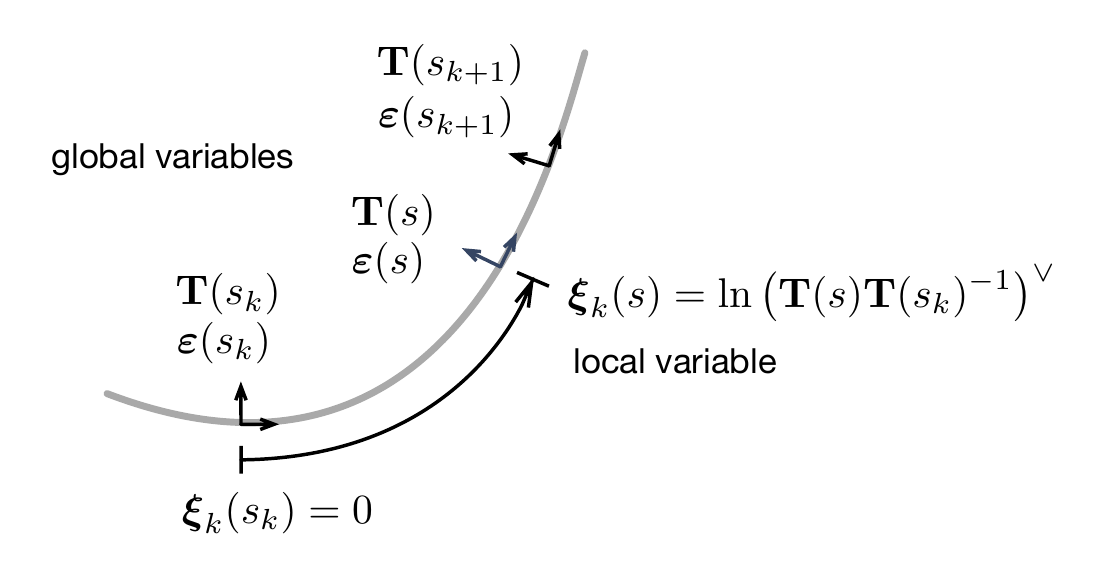}
\caption{Local pose variables, $\mbs{\xi}_k(s)$, are used to simplify the process of defining a \ac{GP} prior over the robot state.}
\label{fig:local}
\end{figure}

The challenge with~\eqref{eq:model} is that while our simplifications have made the differential equation for the physics linear, the kinematics remain nonlinear, which is necessary to allow for large robot displacements.  This makes it fairly difficult to stochastically integrate these models to produce a prior of the form requested in~\eqref{eq:prior}.  To overcome this problem, we will employ the same approach as \citet{anderson_iros15,anderson_phd16} by using a series of local \acp{GP} that are stitched together.  Figure~\ref{fig:local} depicts the idea.  We start by selecting $K$ arclengths, $s_k$, along the robot length at which we initialize each of these local $\acp{GP}$.  We define local pose variables in the Lie algebra, $\mbs{\xi}_k(s) \in \frak{se}(3)$, such that between two arclengths we have
\begin{equation}\label{eq:localvar}
\mbf{T}(s) = \underbrace{\exp\left( \mbs{\xi}_k(s)^\wdg \right)}_{\in \, SE(3)} \mbf{T}(s_k).
\end{equation}
The local variable is essentially compounded onto its associated global pose.  This allows us to convert back and forth between the global pose $\mbf{T}(s)$ and the local pose $\mbs{\xi}_k(s)$.  As long as the rotational motion of this local variable does not become too large, it will be a very good representation of the global pose in its local region.

We now choose to define our \ac{GP} indirectly through a linear, arclength-invariant stochastic differential equation on the local variables (as opposed to the global ones):
\begin{equation}
\frac{d^2}{ds^2} \mbs{\xi}_k(s) = \mbf{w}_k(s), \qquad \mbf{w}_k(s) \sim \mathcal{GP}( \mbf{0}, \mbf{Q}_c(s-s^\prime)).
\end{equation}
In other words, we corrupt the second derivative of pose with a zero-mean, white-noise Gaussian process.  We reorganize this into a first-order stochastic differential equation as
\begin{equation}\label{eq:locallti}
\frac{d}{ds} \bbm \mbs{\xi}_k(s) \\  \mbs{\psi}_k(s) \ebm = \bbm \mbf{0} & \mbf{1} \\ \mbf{0} & \mbf{0} \ebm \underbrace{\bbm \mbs{\xi}_k(s) \\  \mbs{\psi}_k(s) \ebm}_{\mbs{\gamma}_k(s)} + \bbm \mbf{0} \\ \mbf{1} \ebm \mbf{w}_k(s),
\end{equation}
where $\mbs{\psi}_k(s) = \frac{d}{ds}\mbs{\xi}_k(s)$ and $\mbf{1}$ is the identity matrix; while $\mbs{\psi}_k(s)$ is the derivative of local pose, it is not exactly the same as the generalized strain, $\mbs{\varepsilon}(s)$; this will be discussed further below.  Next, we need to (stochastically) integrate this equation once in order to calculate a \ac{GP} for the Markovian state, $\mbs{\gamma}_k(s)$.  

The local variable equation~\eqref{eq:locallti} is linear and stochastic integration can be done in closed form \citep[\S 3.4]{barfoot_ser17}; this was in fact the point of switching to the local variables:
\begin{equation}\label{eq:GP}
\mbs{\gamma}_k(s) \sim \mathcal{GP} \bigl( \underbrace{\mbs{\Phi}(s,s_k) \pri{\mbs{\gamma}}_k(s_k)}_{\rm mean~function},  \underbrace{\mbs{\Phi}(s,s_k) \pri{\mbf{P}}(s_k)  \mbs{\Phi}(s,s_k)^T + \mbf{Q}(s-s_k)}_{\rm covariance~function} \bigr),
\end{equation}
where $\mbs{\Phi}(s,s^\prime)$ is the {\em transition function},
\begin{equation} 
\mbs{\Phi}(s,s^\prime) = \bbm \mbf{1} & (s-s^\prime) \mbf{1} \\ \mbf{0} & \mbf{1} \ebm, \qquad s \geq s^\prime,
\end{equation}
$\mbf{Q}(s-s^\prime)$ is the covariance accumulated between two arclengths,
\begin{equation}
\mbf{Q}(s-s^\prime) = \bbm \frac{1}{3} (s-s^\prime)^3 \mbf{Q}_c & \frac{1}{2} (s-s^\prime)^2 \mbf{Q}_c \\ \frac{1}{2} (s-s^\prime)^2 \mbf{Q}_c & (s-s^\prime) \mbf{Q}_c \ebm, \qquad s \geq s^\prime,
\end{equation}
and $\pri{\mbs{\gamma}}_k(s_k)$ and $\pri{\mbf{P}}(s_k)$ are the initial mean covariance at $s = s_k$, the starting point of the local variable.

\begin{figure}[t]
\centering
\includegraphics[width=0.7\textwidth]{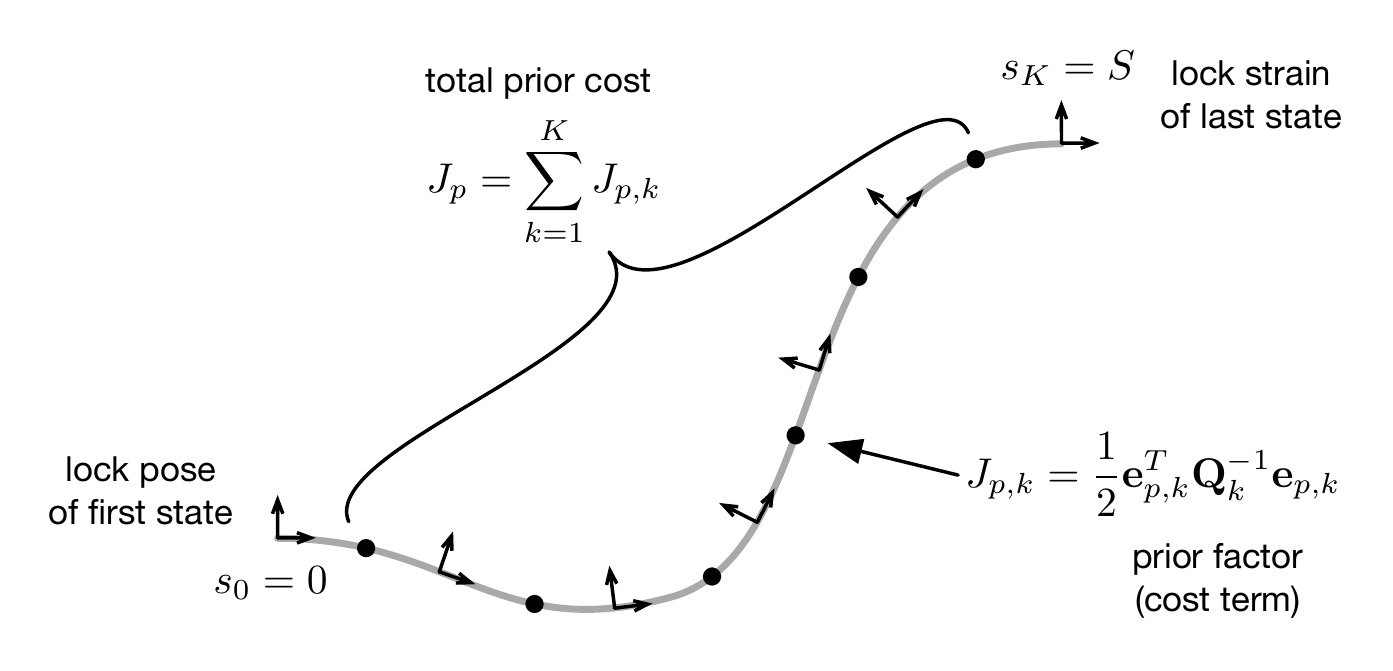}
\caption{The prior for the entire robot state can be broken down into a sequence of binary {\em factors} (black dots), each of which represents a squared error term for how that pair of states is related.  High cost is associated with shapes and strains that deviate from the prior mean (e.g., straight rod with no strain).}
\label{fig:prior}
\end{figure}

Our next goal is construct the overall cost function, whose minimizing solution will be the posterior distribution for the robot state.  It will have two main terms, one for the prior and one for the measurements.  We begin by constructing the prior term in this section and continue with the measurement term in the next.  Normally in \ac{GP} regression we would build a {\em kernel matrix} between all pairs of unique states to be estimated (i.e., all pairs of arclength values whose state is to be estimated).  However, because we have built our \ac{GP} prior from a stochastic differential equation, it has an inherently sparse structure deriving from the Markovian property of the equation and we only need to evaluate the kernel between sequential states along the robot length \citep{barfoot_rss14}.  We define the error between two sequential values of the state as
\begin{equation}\label{eq:localerr}
\mbf{e}_{p,k} = \left( \mbs{\gamma}_k(s_{k}) - \pri{\mbs{\gamma}}_k(s_{k})\right) - \mbs{\Phi}(s_{k},s_{k-1}) \left( \mbs{\gamma}_k(s_{k-1}) - \pri{\mbs{\gamma}}_k(s_{k-1})\right).
\end{equation}
We can construct a cost term for this error as
\begin{equation}\label{eq:cost}
J_{p,k} = \frac{1}{2} \mbf{e}_{p,k}^T \mbf{Q}_k^{-1} \mbf{e}_{p,k},
\end{equation}
where $\mbf{Q}_k = \mbf{Q}(s_{k}-s_{k-1})$.This represents the negative log-likelihood of the error, or in other words (the negative log of) how likely the two consecutive states are at this point on the robot length; the further away the two states are from what the prior mean indicates, the less likely, and this will be traded off against any measurements of the state we also have.  Finally, we can sum all the individual errors between consecutive states along the length to create the total prior cost as
\begin{equation}
J_p = \sum_{k=1}^{K} J_{p,k}.
\end{equation}
Figure~\ref{fig:prior} depicts the cost terms as a {\em factor graph}; each black dot represents one of the squared-error cost terms and the sum of these is our overall prior cost term.

The last detail we need to work out is how to swap out the local variables in the error~\eqref{eq:localerr} for the original global variables, since these are the ones we actually want to estimate.  We can isolate for $\mbs{\xi}_k(s)$ in~\eqref{eq:localvar} so that
\begin{equation}\label{eq:localvar2}
\mbs{\xi}_k(s) = \ln \left( \mbf{T}(s) \mbf{T}(s_k)^{-1} \right)^\vee,
\end{equation}
where $\ln( \cdot)$ is the matrix logarithm and $\vee$ undoes $\wdg$.  We can also write the kinematics of~\eqref{eq:kinematics} using the local variable as
\begin{equation}\label{eq:localkin}
\mbs{\psi}_k(s) = \frac{d}{ds}\mbs{\xi}_k(s) = \mbs{\mathcal{J}}\left( \mbs{\xi}_k(s) \right)^{-1} \mbs{\varepsilon}(s),
\end{equation}
where $\mbs{\mathcal{J}}$ is the left Jacobian of $SE(3)$ \citep[p.236]{barfoot_ser17}.  Inserting~\eqref{eq:localvar2} and~\eqref{eq:localkin} into the error~\eqref{eq:localerr}, we can rewrite it as \citep{anderson_iros15}
\begin{equation}\label{eq:globalerr}
\mbf{e}_{p,k} = \bbm  \ln \left( \mbf{T}(s_k) \mbf{T}(s_{k-1})^{-1} \right)^\vee - (s_k - s_{k-1}) \, \mbs{\varepsilon}(s_{k-1}) \\ \mbs{\mathcal{J}}\left(  \ln \left( \mbf{T}(s_k) \mbf{T}(s_{k-1})^{-1} \right)^\vee \right)^{-1} \mbs{\varepsilon}(s_k) - \mbs{\varepsilon}(s_{k-1}) \ebm,
\end{equation}
which is now only in terms of the global variables, $\{ \mbf{T}(s_k), \mbs{\varepsilon}(s_k) \}$.  

\begin{figure}[t]
\centering
\includegraphics[width=0.7\textwidth]{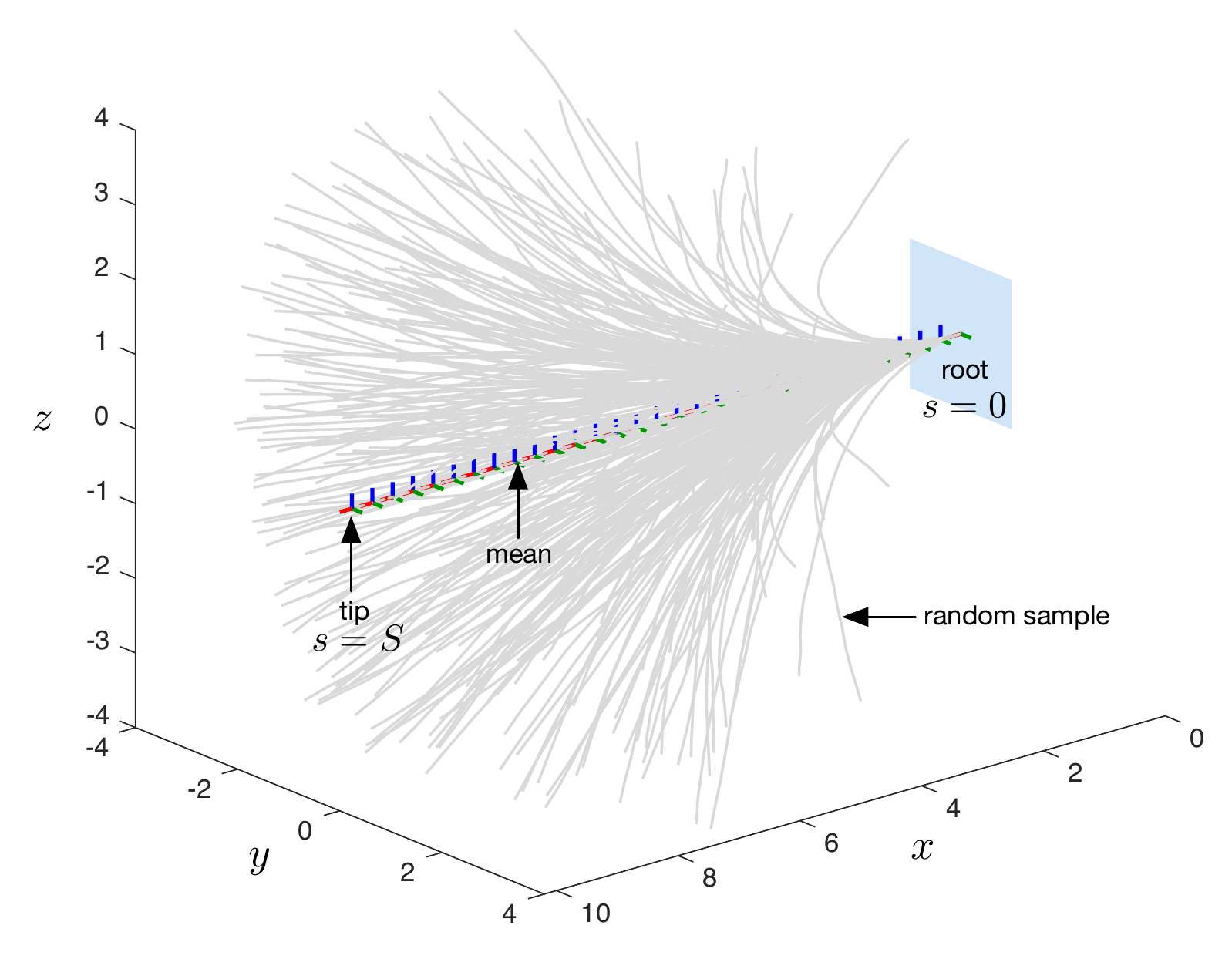}
\caption{Graphical depiction of the prior distribution over robot states.  Here we show the mean shape for the robot, a straight line along the $x$ axis, as well as $300$ random samples drawn from our \ac{GP} prior.   Here we took $S = 10$, $\bar{\mbs{\varepsilon}} = (1,0,0,0,0,0)$, and $\mbf{Q}_c = \mbox{diag}\left(0.01, 0.01, 0.01, 0.001, 0.001, 0.001 \right)$.  In rough terms, our state estimator will downselect these possibilities based on the measurements. }  
\label{fig:samples}
\end{figure}

Another way to think about the development of the prior in this section is that we defined uncertainty affecting our state in the Lie algebra of the pose, where we could avoid the need to worry about the constraints on the variables.  While this development is somewhat involved, the result is quite simple.  We use~\eqref{eq:globalerr} inside~\eqref{eq:cost} to build a bunch of squared-error terms for the states at a number of discrete positions along the arclength of the robot, also known as {\em factors} as depicted in Figure~\ref{fig:prior}.  Figure~\ref{fig:samples} shows the mean function and $300$ random samples drawn from the prior; we can see the prior allows for all kinds of quite different robot shapes.  The {\em hyperparameters} of our prior are the nominal strain, $\bar{\mbs{\varepsilon}}$, as well as the power spectral density, $\mbf{Q}_c$, which controls smoothness in six degrees of freedom.
In this context, choosing lower values for $\mbf{Q}_c$ enforces a smoother rod with respect to the corresponding entries (shear, elongation, bending, twisting), which will be closer to the mean of the prior, which is a straight rod in our case.
In contrast, choosing high values will allow more deformation for the corresponding entries, leading to larger deviations between the resulting shape estimate and a straight rod.

The next section will define similar squared-error terms for our measurements, and then we will then perform an optimization to find the best state for the robot as well as its associated uncertainty.

\subsubsection{Measurements}

\begin{figure}[t]
\centering
\includegraphics[width=0.8\textwidth]{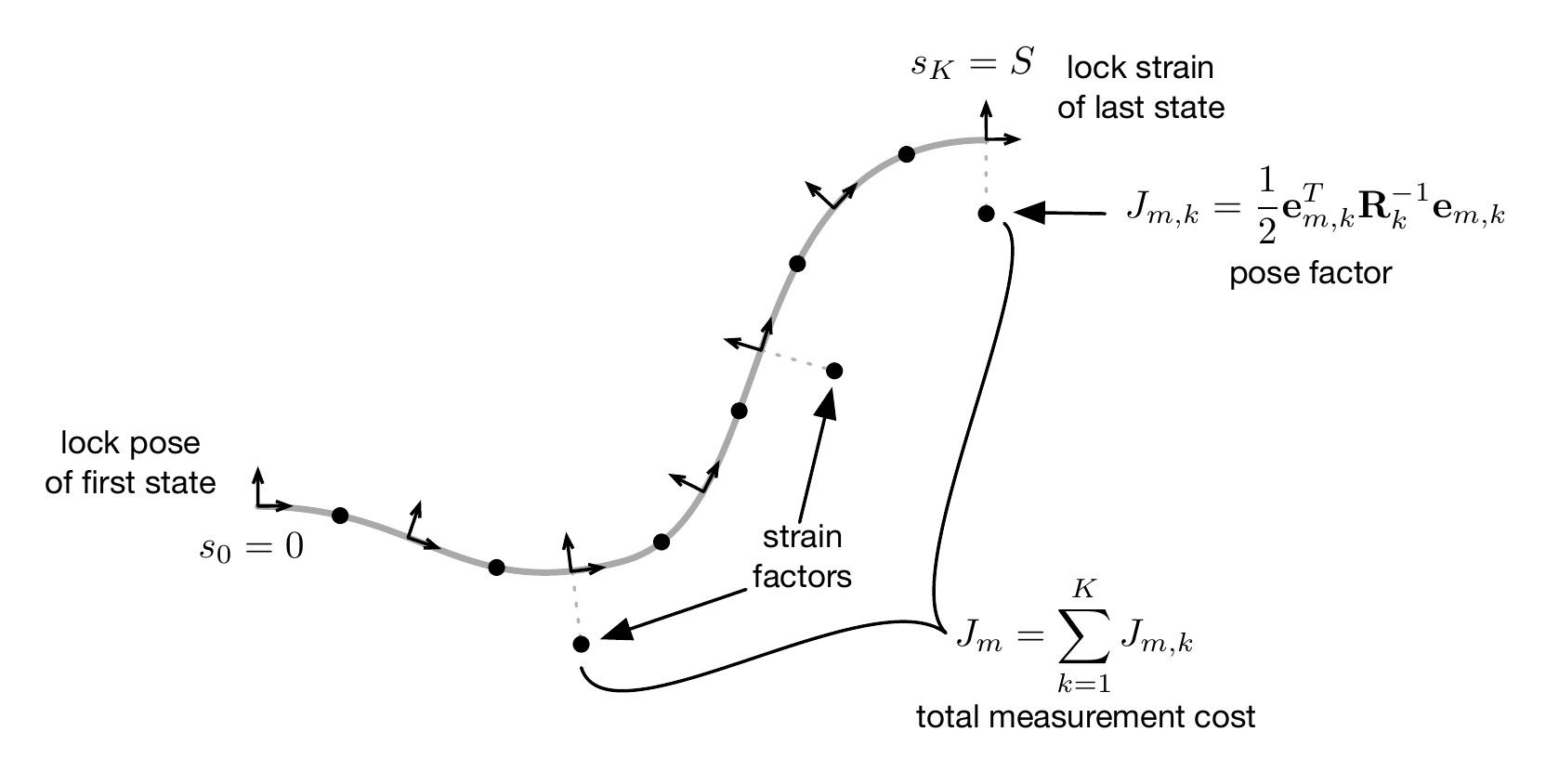}
\caption{The measurements (both pose and strain) make up unary {\em factors} (black dots), each of which represents a squared error term for how that state should be.  High cost is associated with shapes and strains that deviate from the measurements.  Here we have a typical setup with a pose measurement of the end effector and two strain measurements at intermediate arclengths.}
\label{fig:measurements}
\end{figure}

Relatively speaking, the cost terms for the measurements are easier to explain than the prior from the previous section.  We will consider two types of measurements corresponding to commonly used sensing modalities in continuum robotics:
\begin{enumerate}
\item {\em pose measurements:}  the pose (or part thereof) of a continuum robot can be measured at discrete arclengths either by an external camera system or magnetic coils attached to the robot
\item {\em strain measurements:}  the strain (or part thereof) can also be measured at discrete arclengths using embedded strain gauges such as those employing fibre-optical Bragg gratings
\end{enumerate}
We consider that all measurements are noisy and thus we will probabilistically fuse our prior continuum robot model with these measurements to produce a posterior estimate of the shape and strain.  Figure~\ref{fig:measurements} depicts the measurements of pose and strain as {\em unary factors}, meaning squared-error terms involving the state at a single arclength.  The next two subsections lay out the cost terms for the two types of measurements.

\subsubsection{Pose Measurements}

Let us assume that the full pose of the robot state can be measured at a particular arclength, $s_k$.  Following \citet{barfoot_tro14, barfoot_ser17}, we assume this pose measurement, $\widetilde{\mbf{T}}_k$, is drawn from a Gaussian over $SE(3)$:
\begin{equation}
\widetilde{\mbf{T}}_k \leftarrow \mathcal{N} \left( \mbf{T}(s_k), \mbf{R}_k \right),
\end{equation}
where $\mbf{T}(s_k)$ is the true pose and $\mbf{R}_k \in \mathbb{R}^{6\times 6}$ a covariance.  In more detail, what this means is that
\begin{equation}
\widetilde{\mbf{T}}_k  = \exp\left( \mbf{n}_k^\wdg \right) \mbf{T}(s_k)
\end{equation}
where $\mbf{n}_k \in \mathbb{R}^6$ is a regular Gaussian random variable drawn from $\mathcal{N}\left( \mbf{0}, \mbf{R}_k \right)$.

We can form an error for this pose measurement by reorganizing the measurement equation as
\begin{equation}\label{eq:poseerr}
\mbf{e}_{m,k} = \ln \left(  \widetilde{\mbf{T}}_k \mbf{T}(s_k)^{-1} \right)^\vee,
\end{equation}
which will be zero when the pose measurement matches the true state.  The scalar cost term, or factor, is then simply
\begin{equation}\label{eq:posecost}
J_{m,k} = \frac{1}{2} \mbf{e}_{m,k}^T \mbf{R}_k^{-1} \mbf{e}_{m,k},
\end{equation}
which represents the negative log likelihood of the pose measurement.  A projection matrix can be used to mask off unmeasured degrees of freedom in~\eqref{eq:poseerr}, e.g., a motion capture system tracking only position or a magnetic coil unable to measure roll.

\subsubsection{Strain Measurements}

Strain measurements are even simpler to handle than pose measurements as we do not need to contend with $SE(3)$.  We assume a measurement of the strain, $\widetilde{\mbs{\varepsilon}}_k$, is drawn from a Gaussian in $\mathbb{R}^6$:
\begin{equation}
\widetilde{\mbs{\varepsilon}}_k \leftarrow \mathcal{N}\left( \mbs{\varepsilon}(s_k), \mbf{R}_k \right),
\end{equation}
where $\mbs{\varepsilon}(s_k)$ is the true strain at arclength $s_k$ and $\mbf{R}_k \in \mathbb{R}^{6 \times 6}$ a covariance.  We can define the error for this measurement as
\begin{equation}\label{eq:strainerr}
\mbf{e}_{m,k} = \widetilde{\mbs{\varepsilon}}_k -  \mbs{\varepsilon}(s_k),
\end{equation}
and the scalar cost term (factor) as
\begin{equation}\label{eq:straincost}
J_{m,k} = \frac{1}{2} \mbf{e}_{m,k}^T \mbf{R}_k^{-1} \mbf{e}_{m,k},
\end{equation}
which again represents the negative log likelihood of the pose measurement.  If we can only measure particular degrees of freedom of the strain, a projection matrix can easily be used to mask off unmeasured degrees of freedom in~\eqref{eq:strainerr}.

\subsubsection{Optimization}

Putting all of our cost terms together, the overall cost that we seek to minimize is
\begin{equation}\label{eq:totalcost}
J = \sum_{k=1}^K J_{p,k} + \sum_{k=0}^K J_{m,k},
\end{equation}
where we note that not all states will actually have measurements and therefore in practice we include terms only for those measurements actually present.  Note also that the second summation begins at $k=0$ owing to our indexing scheme as depicted in Figures~\ref{fig:prior} and~\ref{fig:measurements}.

Our optimization problem is therefore
\begin{equation}
\est{\mbf{x}} = \mbox{arg}\min_{\mbf{x}} J(\mbf{x}).
\end{equation}
That is, we want the state 
\begin{equation}
\mbf{x} = \bbm \mbf{x}(s_0) \\ \mbf{x}(s_1) \\ \vdots \\ \mbf{x}(s_K) \ebm,
\end{equation}
that minimizes the cost where $\mbf{x}(s_k) = \{\mbf{T}(s_k), \mbs{\varepsilon}(s_k)\}$.  This will be the most likely state for the robot and we will explain how to extract uncertainty information about this estimate below.  Technically, we are abusing notation by writing $\mbf{x}$ as a column like this but it will make the optimization scheme easier to follow; consider it a shorthand.

We use Gauss-Newton optimization to find our estimate, $\hat{\mbf{x}}$, iteratively.  Perturbations to the pose parts of the state will be performed in an $SE(3)$-sensitive way \citep{barfoot_ser17}:
\begin{equation}\label{eq:posepert}
\mbf{T}(s_k) = \underbrace{\exp\left( \delta\mbs{t}_k^\wdg \right)}_{\in \; SE(3)} \mbf{T}(s_k)_{\rm op},
\end{equation}
where $\delta\mbs{t}_k \in \mathbb{R}^6$ is the perturbation and $ \mbf{T}(s_k)_{\rm op}$ the operating point (i.e., value of the estimate from the previous iteration).  Perturbations to the strain parts of the state will be performed in the usual vectorspace way:
\begin{equation}\label{eq:strainpert}
\mbs{\varepsilon}(s_k) = \mbs{\varepsilon}(s_k)_{\rm op} + \delta\mbs{\varepsilon}_k,
\end{equation}
where $\delta\mbs{\varepsilon}_k \in \mathbb{R}^6$ is the perturbation and $\mbs{\varepsilon}(s_k)_{\rm op}$ the operating point.  We can assemble the two parts of the state together and write
\begin{equation}
\mbf{x}(s_k) = \mbf{x}(s_k)_{\rm op} + \delta \mbf{x}_k, \qquad \delta \mbf{x}_k = \bbm \delta\mbs{t}_k \\ \delta\mbs{\varepsilon}_k \ebm,
\end{equation}
or 
\begin{equation}
\mbf{x} = \mbf{x}_{\rm op} + \delta \mbf{x}, \qquad \delta\mbf{x} = \bbm \delta\mbf{x}(s_0) \\ \vdots \\ \delta\mbf{x}(s_K) \ebm,
\end{equation}
for the entire robot state.

We can rewrite our cost function~\eqref{eq:totalcost} using stacked quantities rather than summations as
\begin{equation}\label{eq:totalcost2}
J = \frac{1}{2} \mbf{e}_p^T \mbf{Q}^{-1} \mbf{e}_p + \frac{1}{2} \mbf{e}_m^T \mbf{R}^{-1} \mbf{e}_m,
\end{equation}
where
\begin{equation}
\mbf{e}_p = \bbm \mbf{e}_{p,1} \\ \vdots \\ \mbf{e}_{p,K} \ebm, \quad \mbf{Q} = \mbox{diag}\left( \mbf{Q}_1, \ldots, \mbf{Q}_K \right), \quad \mbf{e}_m = \bbm \mbf{e}_{m,0} \\ \vdots \\ \mbf{e}_{m,K} \ebm, \quad \mbf{R} = \mbox{diag}\left( \mbf{R}_0, \ldots, \mbf{R}_K \right).
\end{equation}
Inserting our perturbation schemes from~\eqref{eq:posepert} and~\eqref{eq:strainpert} we can linearize the prior and measurement errors as
\begin{equation}\label{eq:linerrors}
\mbf{e}_p = \mbf{e}_{p,{\rm op}} + \mbf{E}_p \, \delta\mbf{x}, \qquad \mbf{e}_m = \mbf{e}_{m,{\rm op}} + \mbf{E}_m \, \delta\mbf{x},
\end{equation}
where $\mbf{e}_{p,{\rm op}}$ and $\mbf{e}_{m,{\rm op}}$ are the errors evaluated at the previous iteration's state values and $\mbf{E}_p$ and $\mbf{E}_m$ are Jacobians (details to follow in next section).  Inserting~\eqref{eq:linerrors} into~\eqref{eq:totalcost2} we have
\begin{equation}
J \approx \frac{1}{2} \left( \mbf{e}_{p,{\rm op}} + \mbf{E}_p \, \delta\mbf{x} \right)^T \mbf{Q}^{-1} \left( \mbf{e}_{p,{\rm op}} + \mbf{E}_p \, \delta\mbf{x} \right) + \frac{1}{2} \left( \mbf{e}_{m,{\rm op}} + \mbf{E}_m \, \delta\mbf{x} \right)^T \mbf{R}^{-1} \left( \mbf{e}_{m,{\rm op}} + \mbf{E}_m \, \delta\mbf{x} \right),
\end{equation}
which is now quadratic in our perturbation variable, $\delta\mbf{x}$.  The linear system of equations,
\begin{equation}\label{eq:linsys}
\underbrace{\left( \mbf{E}_p^T \mbf{Q}^{-1} \mbf{E}_p + \mbf{E}_m^T \mbf{R}^{-1} \mbf{E}_m \right)}_{\rm block-tridiagonal} \, \delta\mbf{x}^\star = - \left( \mbf{E}_p^T \mbf{Q}^{-1}  \mbf{e}_{p,{\rm op}} + \mbf{E}_m^T \mbf{R}^{-1} \mbf{e}_{m,{\rm op}} \right),
\end{equation}
provides the minimizing solution, $\delta\mbf{x}^\star$.  Owing to the nature of our factors (binary and unary), this is a highly sparse linear system whose solution can be found in $O(K)$ time using, for example, sparse Cholesky decomposition and forward/backward passes \citep{meurant1992review}. Once we have our optimal perturbation for this iteration, we can unstack it into its constituent pieces and update our state variable operating points using our original perturbation schemes:
\beqn{}
\mbf{T}(s_k)_{\rm op} & \leftarrow & \exp\left( \delta\mbs{t}_k^{\star^\wdg} \right) \mbf{T}(s_k)_{\rm op}, \\
\mbs{\varepsilon}(s_k)_{\rm op} & \leftarrow & \mbs{\varepsilon}(s_k)_{\rm op} + \delta\mbs{\varepsilon}_k^\star,
\eeqn
which for the pose part of the state ensures we remain within $SE(3)$ at each iteration.  We iterate this entire scheme to convergence (i.e., $\delta\mbf{x}$ becomes sufficiently small) and then take the values at the final iteration to be our estimates, $\hat{\mbf{x}}(s_k) = \{ \hat{\mbf{T}}(s_k), \hat{\mbs{\varepsilon}}(s_k) \}$.  It is also worth noting that to enforce our boundary conditions, we set $\mbf{T}(s_0) = \mbf{1}$ and $\mbs{\varepsilon}(s_K) = \bar{\mbs{\varepsilon}}$ at the first iteration and do not update them during the optimization.

\begin{figure}[t]
\centering
\includegraphics[width=\textwidth]{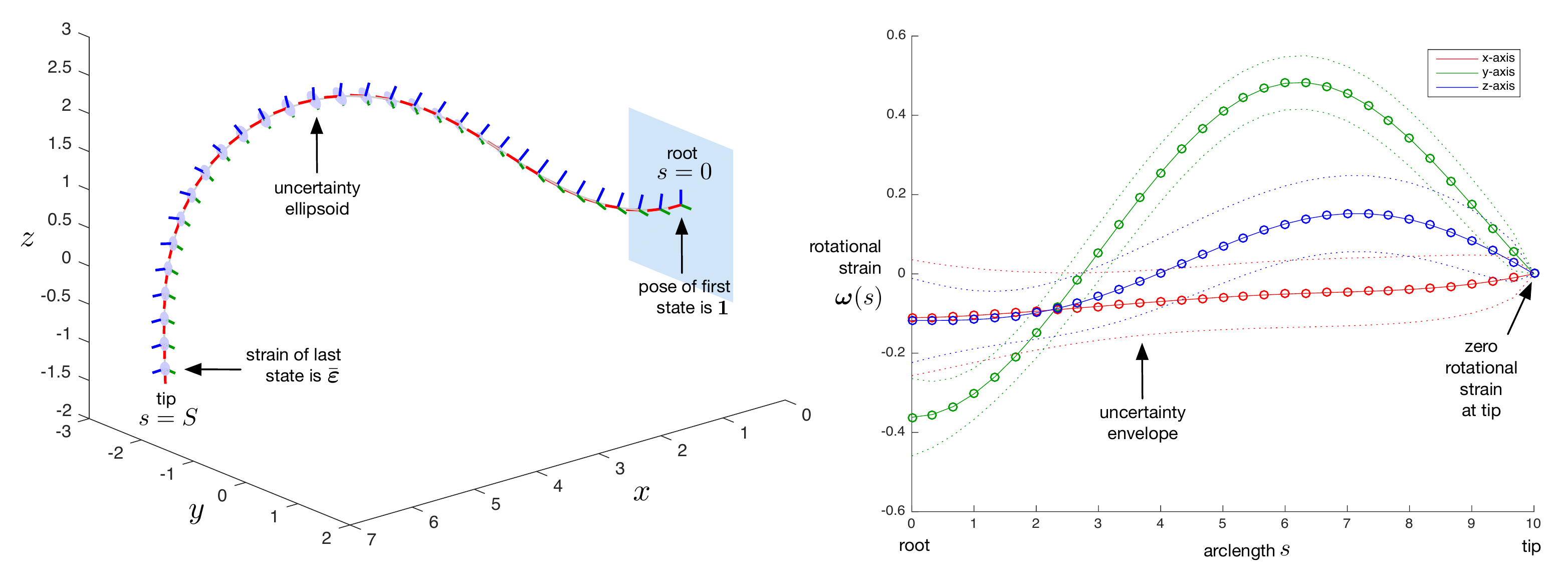}
\caption{Example solution from our continuum robot state estimator with $K=30$.  A single noisy pose measurement of the end effector (located at $(6,-2,-1)$ pointing down) is fused with the kinematics/physics prior.  The left plot depicts the pose part of the state estimate including 3$\sigma$ uncertainty ellipsoids for the position. The pose of the root is constrained to an identity transform and the strain of the tip is constrained to the nominal strain, $\bar{\mbs{\varepsilon}} = (1, 0, 0, 0, 0, 0)$.  Translational strains for all states were also locked to their nominal values (i.e., unshearable and inextensible).  As expected, we see the largest uncertainties in the middle section of the robot.  The right plot shows the rotational part of the strain components along with 3$\sigma$ uncertainty envelopes. This estimate converged in approximately $5$ iterations. }
\label{fig:example}
\end{figure}

Figure~\ref{fig:example} provides an example of the estimator in action.  A single measurement of the end effector pose is fused with the kinematics/physics prior.  The caption provides further details.  

\subsubsection{Error Jacobians}

An outstanding item is to provide the details for the error Jacobians, $\mbf{E}_p$ and $\mbf{E}_m$, required in~\eqref{eq:linsys}.  Using our state-variable perturbation schemes, an individual prior error can be linearized as \citep[p.117]{anderson_phd16},
\begin{equation}
\mbf{e}_{p,k} = \mbf{e}_{p,k,{\rm op}} + \underbrace{\bbm -\mbs{\mathcal{J}}_{k,k-1}^{-1} \mbs{\mathcal{T}}_{k,k-1} & - (s_k - s_{k-1}) \mbf{1} & \mbs{\mathcal{J}}_{k,k-1}^{-1} & \mbf{0} \\ -\frac{1}{2} \mbs{\varepsilon}_{k,{\rm op}}^\Wdg \mbs{\mathcal{J}}_{k,k-1}^{-1} \mbs{\mathcal{T}}_{k,k-1} & - \mbf{1} & \frac{1}{2} \mbs{\varepsilon}_{k,{\rm op}}^\Wdg \mbs{\mathcal{J}}_{k,k-1}^{-1} & \mbs{\mathcal{J}}_{k,k-1}^{-1} \ebm}_{\mbf{E}_{p,k}} \bbm \delta\mbs{t}_{k-1} \\ \delta\mbs{\varepsilon}_{k-1} \\ \delta\mbs{t}_{k} \\ \delta\mbs{\varepsilon}_{k}\ebm,
\end{equation}
where
\begin{equation}
\mbs{\mathcal{J}}_{k,k-1} = \mbs{\mathcal{J}}\left( \ln \left(   \mbf{T}(s_k)_{\rm op} \mbf{T}(s_{k-1})_{\rm op}^{-1} \right)^\vee  \right), \qquad \mbs{\mathcal{T}}_{k,k-1} = \mbox{Ad}\left( \mbf{T}(s_k)_{\rm op} \mbf{T}(s_{k-1})_{\rm op}^{-1} \right).
\end{equation}
The blocks of $\mbf{E}_{p,k}$ must be placed into the appropriate entries of the overall Jacobian, $\mbf{E}_p$, according to the indices of the variables involved.

For the pose measurements, the error can be linearized as
\begin{equation}
\mbf{e}_{m,k} = \mbf{e}_{m,k,{\rm op}} + \underbrace{\bbm -\mbs{\mathcal{J}}\left(  \ln \left(   \widetilde{\mbf{T}}_k \mbf{T}(s_{k})_{\rm op}^{-1} \right)^\vee \right)^{-1} \mbox{Ad}\left(\widetilde{\mbf{T}}_k \mbf{T}(s_k)_{\rm op}^{-1}\right) & \mbf{0} \ebm}_{\mbf{E}_{m,k}} \bbm \delta\mbs{t}_k \\ \delta \mbs{\varepsilon}_k \ebm.
\end{equation}
For the strain measurements, the error linearizes simply as
\begin{equation}
\mbf{e}_{m,k} = \mbf{e}_{m,k,{\rm op}} + \underbrace{\bbm \mbf{0} & - \mbf{1} \ebm}_{\mbf{E}_{m,k}} \bbm \delta\mbs{t}_k \\ \delta \mbs{\varepsilon}_k \ebm.
\end{equation}
As with the prior terms, the blocks of $\mbf{E}_{m,k}$ (whether from pose or strain) must be placed into the appropriate blocks of the overall Jacobian, $\mbf{E}_m$, according to the variables involved.

\begin{figure}[t]
\centering
\includegraphics[width=\textwidth]{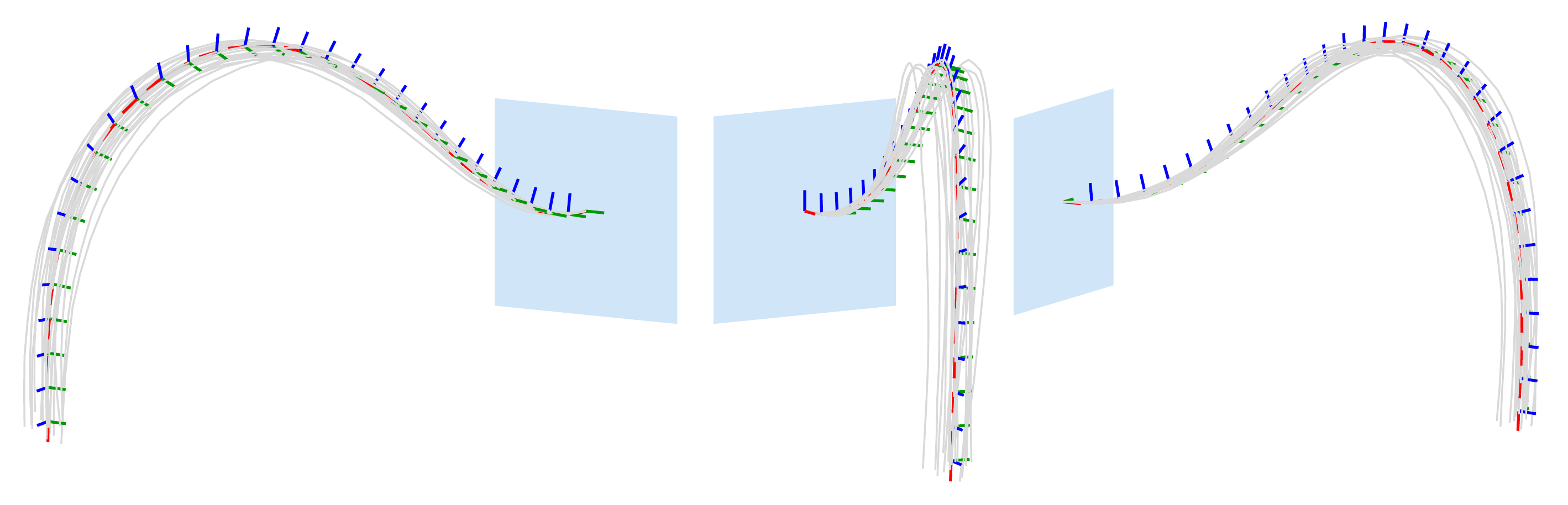}
\caption{Three viewpoints of an estimate with its mean (red-green-blue frames) and $20$ samples drawn from the full Gaussian distribution (gray lines).}
\label{fig:posterior}
\end{figure}

\subsubsection{Extracting Uncertainties}

We have so far only discussed how to determine the most likely robot state in our optimization setup.  To extract the uncertainties of the estimator (e.g., see Figure~\ref{fig:example}) we can turn to the common Laplace approximation, which fits a covariance at the most likely state.  Naively, we could take the left-hand side of the linear system of equations~\eqref{eq:linsys} at the last iteration and invert this for the covariance, $\hat{\mbf{P}}$:
\begin{equation}\label{eq:fullcov}
\hat{\mbf{P}} = \left(  \mbf{E}_p^T \mbf{Q}^{-1} \mbf{E}_p + \mbf{E}_m^T \mbf{R}^{-1} \mbf{E}_m \right)^{-1}.
\end{equation}
Although $\hat{\mbf{P}}^{-1}$ is block-tridiagonal, $\hat{\mbf{P}}$ will be dense.  We may, however, only require the $12 \times 12$ diagonal blocks of $\hat{\mbf{P}}$ corresponding to each arclength at which we estimated the state.  In this case, we can piggyback the solution of just these blocks onto the solution of the linear system~\eqref{eq:linsys} using the \ac{RTS} smoother \citep{rauch65}, albeit with arclength rather than time.

It is worth mentioning how to interpret the $12 \times 12$ diagonal covariance block $\hat{\mbf{P}}_k$ associated with $\mbf{x}(s_k) = \{ \mbf{T}(s_k), \mbs{\varepsilon}(s_k) \}$.  We can further break this down into subblocks:
\begin{equation}
\hat{\mbf{P}}_k = \bbm \hat{\mbf{P}}_{11,k} & \hat{\mbf{P}}_{12,k} \\ \hat{\mbf{P}}_{12,k}^T & \hat{\mbf{P}}_{22,k} \ebm,
\end{equation} 
where the `1' partition corresponds to pose and the `2' partition corresponds to strain.  Our marginal posterior estimate for the pose is thus
\begin{equation}
\mbf{T}(s_k) \sim \mathcal{N}\left( \hat{\mbf{T}}(s_k), \hat{\mbf{P}}_{11,k} \right),
\end{equation}
which we must interpret in the proper $SE(3)$ way to mean
\begin{equation}
\mbf{T}(s_k) = \exp\left( \mbf{n}_k^\wdg \right) \hat{\mbf{T}}_k, \qquad \mbf{n}_k \sim \mathcal{N}\left( \mbf{0}, \hat{\mbf{P}}_{11,k} \right).
\end{equation}
In other words, the uncertainty is expressed in the Lie algebra associated with the pose.  Our marginal posterior estimate for the strain is simply
\begin{equation}
\mbs{\varepsilon}(s_k) \sim \mathcal{N} \left( \hat{\mbs{\varepsilon}}(s_k), \mbf{P}_{22,k} \right),
\end{equation}
where we do not need to worry about $SE(3)$ since $\mbs{\varepsilon}(s_k) \in \mathbb{R}^6$.

Another nice property of having set up the problem as a batch estimation, is that we can in fact draw samples for the entire robot state from the full covariance in~\eqref{eq:fullcov}.  Figure~\ref{fig:posterior} shows an example where we have drawn $20$ samples from the posterior and displayed them along with the mean.  This is only possible if we have the full covariance since all the states are correlated with one another.

\begin{figure}[t]
\centering
\includegraphics[width=\textwidth]{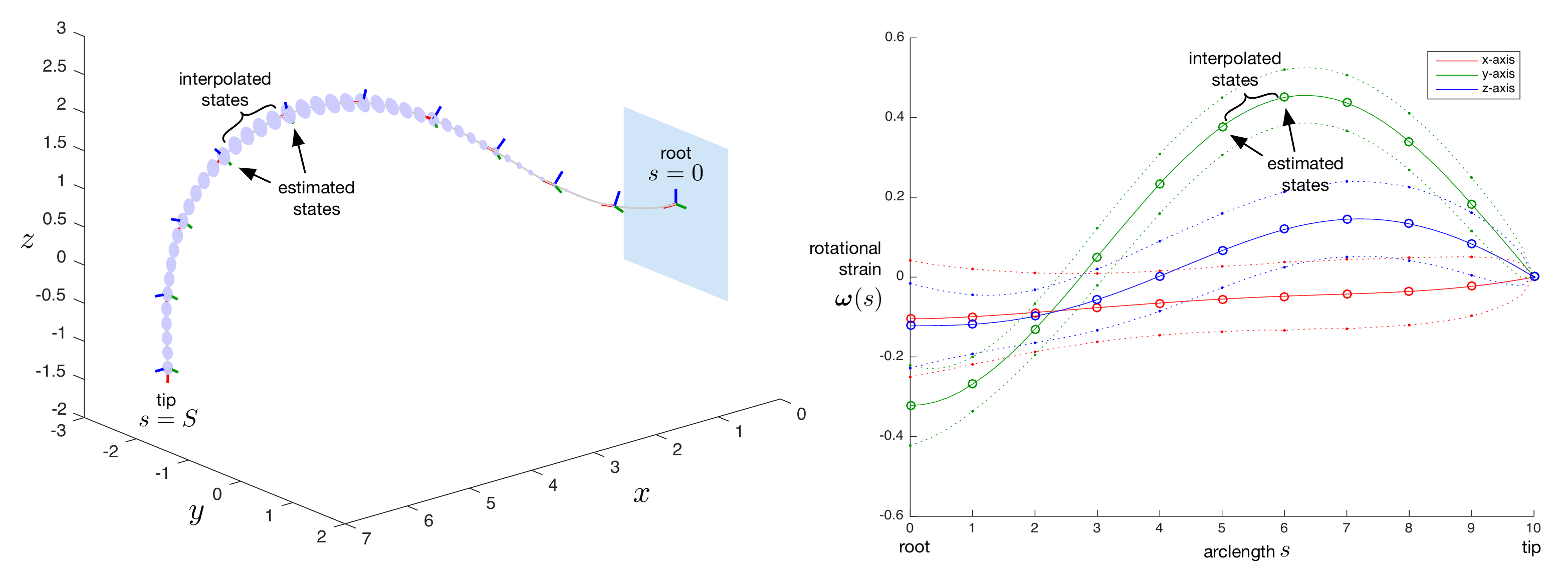}
\caption{One of  appealing features of using the \ac{GP} approach is the ability to query the state at any arclength of interest, after the main estimation has converged.  In this example, we used $K=10$ for the original estimation problem and then interpolated $4$ extra state values between the estimated states.  We can see that this interpolation is not linear but rather that it is based on the prior model we selected allowing it to be curved.  Note, both the mean and the covariance can be queried in this way.}
\label{fig:interp}
\end{figure}

\subsubsection{Querying the State Between Estimated Arclengths}

Because we have set up our problem using \ac{GP} tools, we can easily use the built-in \ac{GP} interpolation equations to query the mean and covariance of the state at any arclength of interest, including at arclengths not computed in the main optimization.  For example, we might want to check how close our robot is to obstacles in a known environment, and we might therefore like to check the pose at a large number of arclengths (e.g., $10$ times as many as were estimated) to be very confident a small object or corner does not come too close.

As shown by \citet{barfoot_rss14,anderson_iros15,barfoot_ser17}, we can perform a state (and covariance) query in $O(1)$ time due to the special nature of our prior (i.e., that it came from a differential equation).  If we want to make our solver as fast as possible, therefore, we would make the number of arclengths in the main solution, $K$, as small as possible and then query additional states after the fact.  We omit the details in the interests of space but refer to \citet{anderson_iros15,anderson_phd16}.

Figure~\ref{fig:interp} shows an example where we used only $K=10$ for the original estimation problem and then queried $4$ additional state values between each pair of estimated states.  Importantly, we can see that the interpolation is not linear with respect to arclength, but allows the robot to be curved.  As discussed by \citet[p.86]{barfoot_ser17}, the particular prior model that we chose (i.e., white noise on strain rate) results in cubic Hermite polynomial interpolation for the robot poses; however, this is a consequence of the prior selection and we can use the general \ac{GP} interpolation equations without worrying about these details. 

\section{Evaluation}

The state estimation approach proposed in this work is evaluated both in simulations and through experiments.
While the proposed algorithm works for any structure or manipulator that can be accurately modeled using Cosserat theory of elastic rods, we focus on continuum robots actuated via tendons throughout this section as an example, which is one of the most commonly used and studied continuum robot types.
This section first introduces the main principle of such so-called tendon-driven continuum robots.
Afterwards, experiments to evaluate the proposed approach in both simulations and on a real robotic prototype are described, before presenting and discussing the achieved results.

\subsection{Tendon-Driven Continuum Robots}

Tendon actuation is one of the most commonly studied actuation principles for continuum robots to date.
These so-called tendon-driven continuum robots (TDCR) are realized by multiple tendons that are routed along their flexible backbone.
The tendons terminate at predefined locations along the robot's arclength, $s$.
Thus, a TDCR can consist of multiple segments, with each segment end being defined by the termination of one of multiple tendons.
Pulling and releasing these tendons applies a load to the compliant backbone, bending the corresponding segment in the direction of the routed tendon.
Thus, the configuration $\mbf{q} = \begin{bmatrix}
\tau_1 & \cdots & \tau_n
\end{bmatrix}^T$ of a TDCR is generally defined as the tension applied to each routed tendon $i \in \left\{1,...,n\right\}$.
We refer to \cite{rao2021model} for a more thorough introduction to TDCR with respect to their mechanical design, actuation principles as well as modeling approaches.

Throughout this work, the proposed state estimation approach is evaluated on a TDCR consisting of two bending segments.
For actuation, tendons, that are routed straight and parallel to the backbone, are considered for each individual segment.
The two bending segments of the TDCR are identical and their parameters are stated in Table~\ref{tab:tdcr_params}.
These parameters include the backbone's Young's Modulus, $E$, its diameter, $d$, the pitch radius, $r$, which describes the distance between the routed tendons and the backbone, as well es the individual segment's length, $L$.

\begin{table}[ht!]\renewcommand*{\arraystretch}{1.5}
	\centering
	\caption{TDCR segment paramaters}
	\begin{tabular}{c c c c c}
		\hline Young's Modulus $E$ & Backbone Diameter $d$ & Pitch Radius $r$ & Length $L$ & Number of Disks \\
		\hline    \hline
		54 GPa & 1 mm & 7 mm & 0.14 m & 7 \\
		\hline
	\end{tabular}
	\label{tab:tdcr_params}
\end{table}

\subsection{Simulation}

For simulation, a TDCR is simulated using the state-of-the-art Cosserat Rod-based static model proposed by \cite{rucker2011statics}.
In particular, we are using a C++ implementation of that model published by \cite{rao2021model}.
An example rendering of the simulated two-segment TDCR is shown in Figure~\ref{fig:simulation_samples} (left).

\subsubsection{Data Set Generation}

\begin{figure}[ht!]
	\centering
	\includegraphics[width=0.325\textwidth]{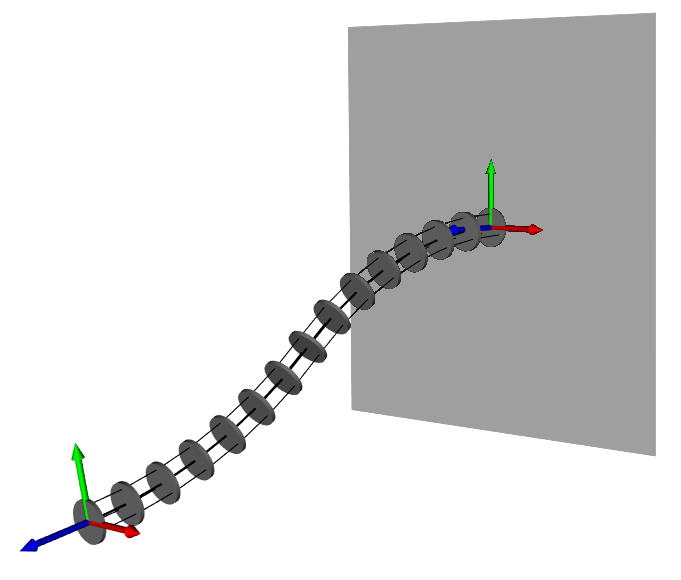}
	\hspace{2cm}
	\includegraphics[width=0.375\textwidth]{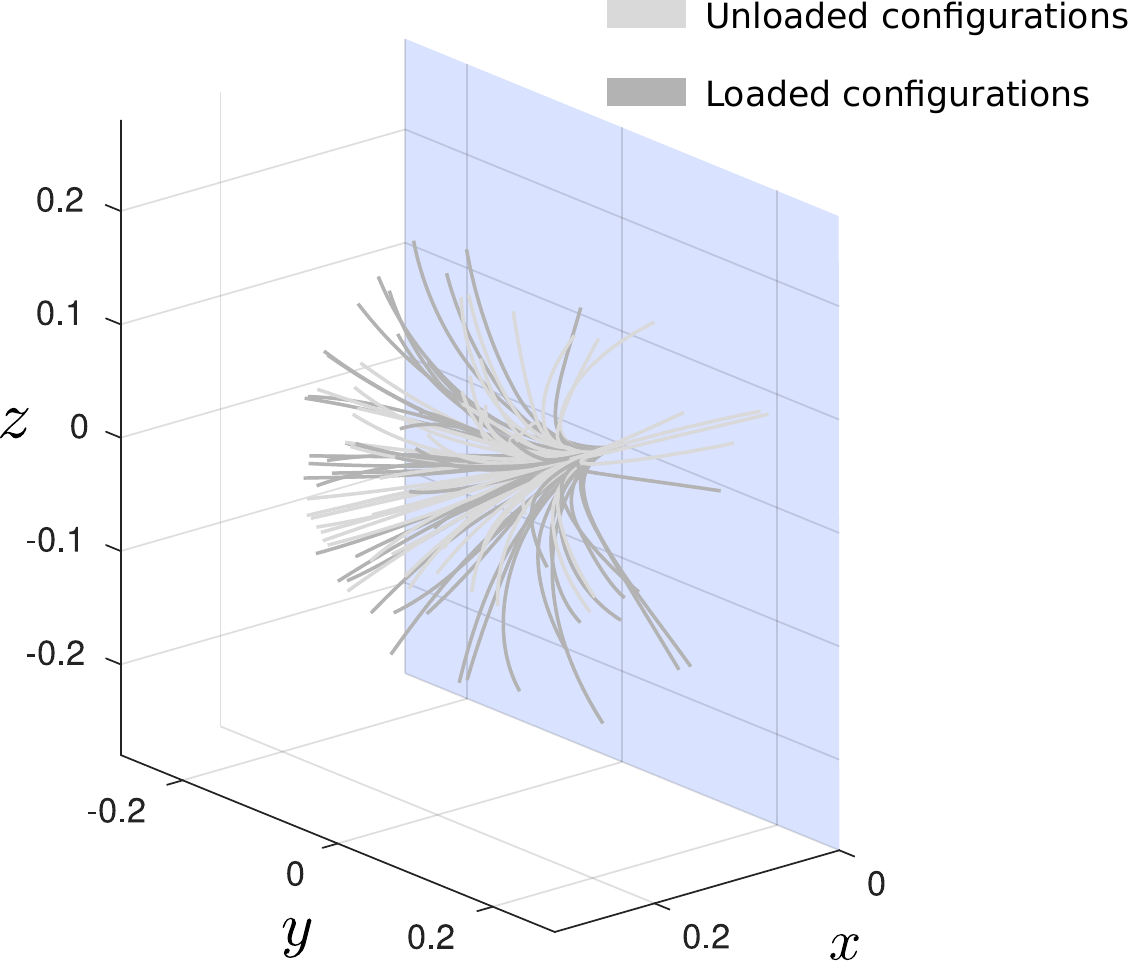}
	\caption{Left: Example rendering of a simulated tendon-driven continuum robot consisting of two bending segments. Right: Overview of 100 random samples of the simulated TDCR. Half of the samples are unloaded configurations, meaning that the deformation of the robot solely resulted from pulling different tendons (light gray), while the other half includes a randomly generated wrench acting at the tip of the robot (dark gray).}
	\label{fig:simulation_samples}
\end{figure}

Using the simulated TDCR, a data set consisting of 100 different robot configuration is sampled.
This is done by drawing random samples from the valid joint space of the robot and using the forward kinematics to obtain the resulting robot state.
Specifically, we consider tendon forces of up to 3N, so that each randomly sampled joint value is $\tau_i \in \left[0 \mathrm{N},3 \mathrm{N}\right]$, while also respecting joint space constraints (usually only up to two tendons are pulled simultaneously).
The sampled robot state includes both the robot's pose $\mbf{T}(s)$ and generalized strain $\mbs{\varepsilon}(s)$.
While half of the considered configurations are load-free, meaning the deformation of the robot solely results from tendon actuation, we considered an additional load acting at the tip of the robot for the remaining configurations.
This load consists of both a randomly sampled force and moment, where each force component is sampled between $\left[-0.1\mathrm{N},0.1\mathrm{N}\right]$ and each moment component is sampled between $\left[-0.01\mathrm{Nm},0.01\mathrm{Nm}\right]$, respectively.
The resulting robot shapes of all of the configurations contained in the data set are plotted in Figure~\ref{fig:simulation_samples} (right).
The unloaded configurations are displayed in light gray, while the configurations subject to an additional load are plotted in dark gray.
This data set serves as ground truth data for the experiments conducted in simulation.

\subsubsection{Creating Noisy Sensor Data}

\begin{figure}[ht!]
	\centering
	\includegraphics[width=1\textwidth]{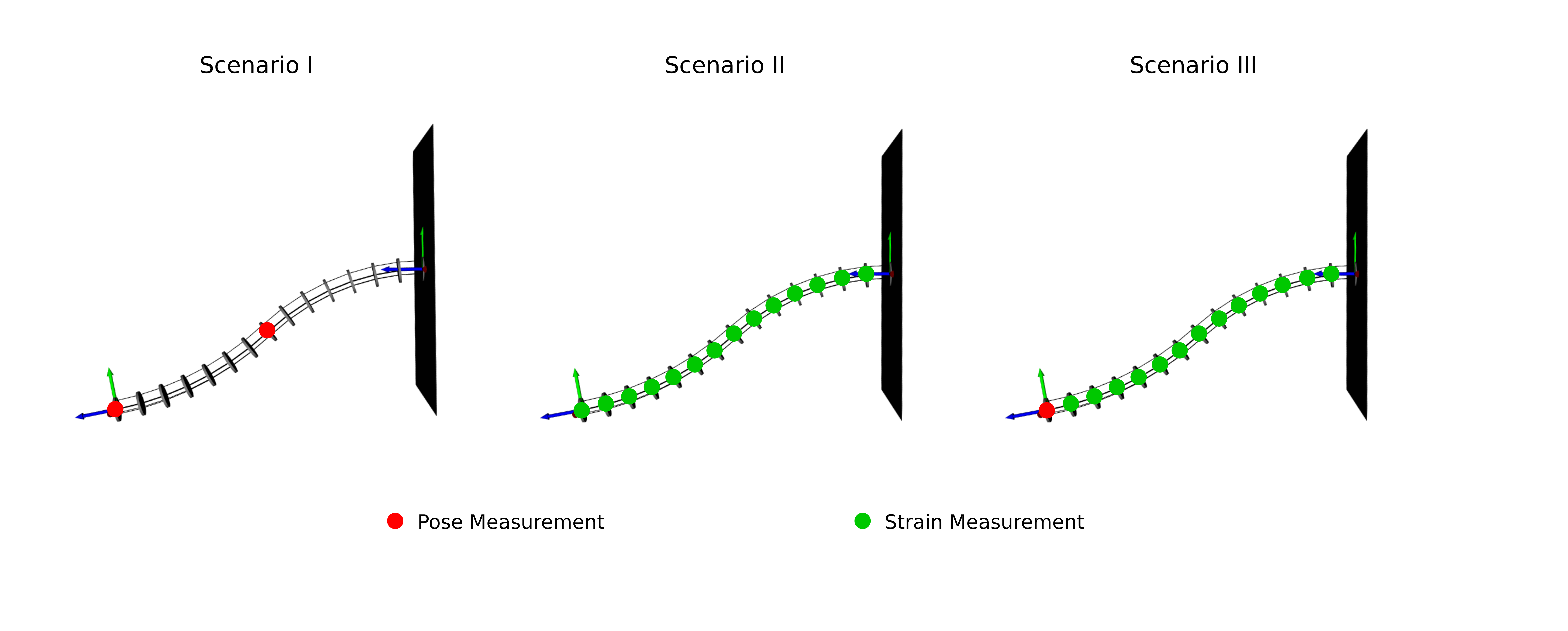}
	\caption{To evaluate the state estimation algorithm proposed in this work, three different sensor placement scenarios are considered: pose measurements at the end of each robot segment (left), discrete strain measurements at each robot disk (middle), discrete strain measurements at each disk in combination with a single pose measurement at the robot's tip (right).}
	\label{fig:sensor_placements}
\end{figure}

In the next step, noisy sensor measurements are simulated using the ground truth data set.
We consider three different scenarios for possible sensor placements (see Figure~\ref{fig:sensor_placements}), each of which are commonly used and proposed for continuum robots:
\begin{enumerate}
	\item Pose measurements at the end of each robot segment (i.e., at its tip and at half of its length)
	\item Multiple discrete strain measurements at each of the robot's disks' location
	\item Multiple discrete strain measurements in combination with a single pose measurement at the robot's tip
\end{enumerate}
Both, pose and strain information are relatively common when sensing the state of a continuum robot.
For instance, an accurate pose measurement can be obtained using electromagnetic tracking coils \cite{Mahoney2016}, while strain measurements can be enabled using Fiber Bragg Gratings (\cite{Shi2017}).

We simulate the sensor measurements by extracting the pose and strain values at the corresponding arclengths from our simulated ground truth data set.
Afterwards, we inject noise on each of the components of the simulated measurements.
This is done by drawing random values from a zero-mean normal distribution, where standard deviation is the noise variable.
These noise variables are chosen in a way to mimic realistic conditions and errors of the particular sensors that would be used to obtain these measurements.
In particular, we chose the standard deviations for the position and orientation of the pose as $\sigma_t = 1~\mathrm{mm}$ and $\sigma_a = 0.01~\mathrm{rad}$, respectively.
The standard deviations for both the translational and rotational strain are set to $\sigma_\nu = \sigma_\om = 0.05$.
While for the strain variables the noise can simply be added to the vector itself, we inject the noise to the pose in an $SE(3)$-sensitive way utilizing Lie algebra, cf. equation \eqref{eq:posepert}.
Afterwards, we can use the matrix exponential to map it to a transformation matrix and apply it to update and displace the pose measurement.

\subsubsection{Hyperparameter Tuning}

\label{sec:hyperparameter}

\begin{figure}[ht!]
	\centering
	\includegraphics[width=1\textwidth]{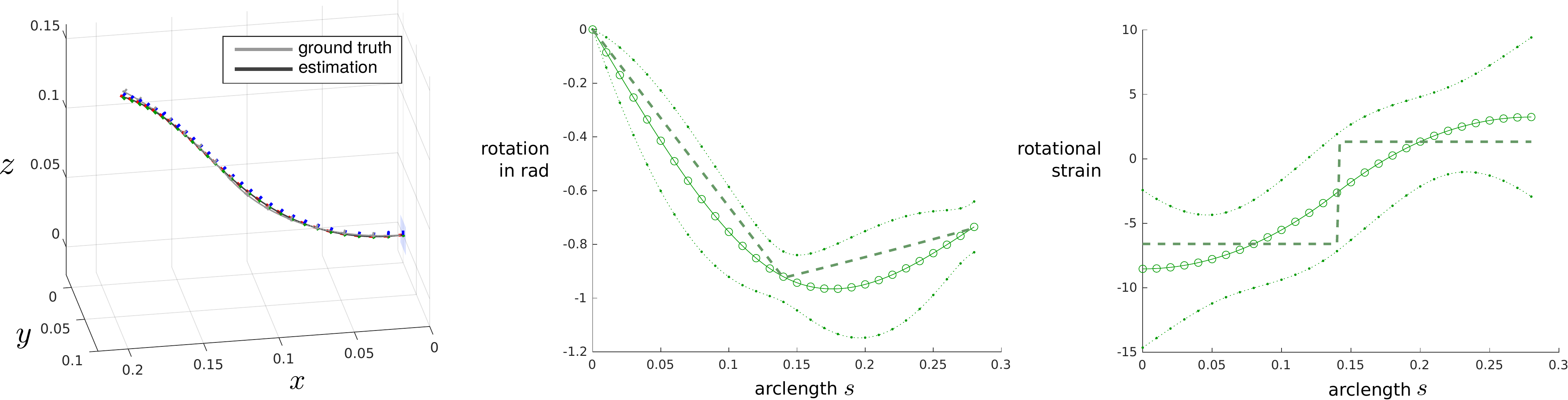}
	\caption{Example TDCR configuration, in which the deformation is mainly dominated by bending around the y-axis. The plots on the left show the $y$-component of the robot's orientation (expressed as an angle-axis pair) as well as the $y$-component of the rotational strain along its arclength $s$. Each plot shows the mean along with the 3$\sigma$ uncertainty envelopes and the simulated ground truth (dashed line).}
	\label{fig:tdcr_strain_example}
\end{figure}

\begin{figure}[b!]
	\centering
	\includegraphics[width=1\textwidth]{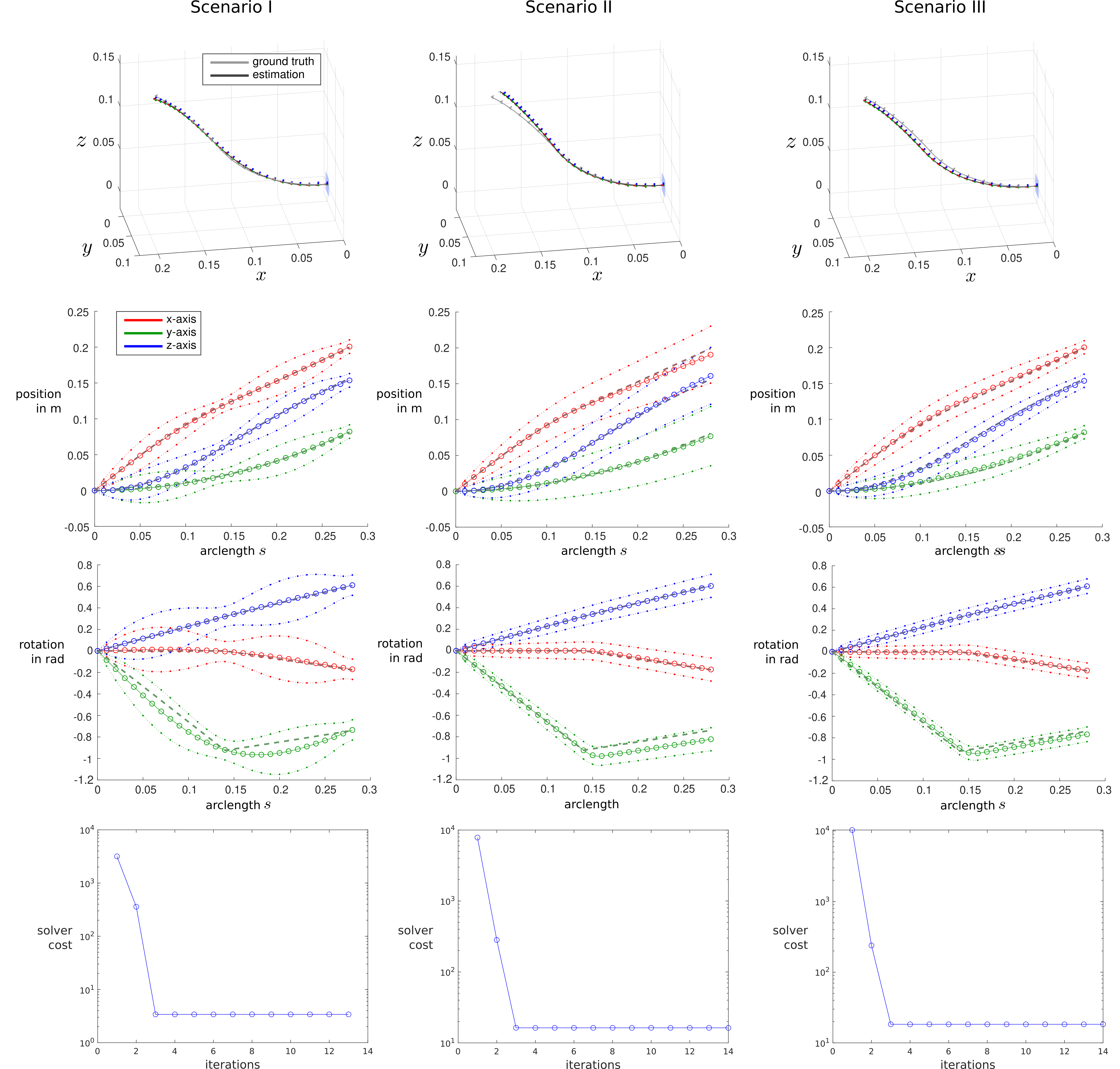}
	\caption{Results from the proposed state estimation algorithm using a typical TDCR configuration. Three different sensor placement scenarios are considered: Pose measurements at the end of each robot segment (left), discrete strain measurements at each robot disk (middle), discrete strain measurements at each disk in combination with a single pose measurement at the robot's tip (right). The resulting continuous estimated shapes are plotted using colored frames on the left with ground truth shapes plotted in gray. Additional plots display the $x$- (red), $y$- (green) and $z$-components (blue) of the robot's translation and orientation (expressed as angle-axis pair) along its arclength $s$. The plots show the mean along with the 3$\sigma$ uncertainty envelopes and the simulated ground truth data (dashed line). The bottom plots show the cost of the optimization problem over the number of iterations.}
	\label{fig:simulation_example}
\end{figure}

All hyperparameters were tuned empirically using a typical TDCR configuration (see Figure~\ref{fig:tdcr_strain_example} and~\ref{fig:simulation_example}) and considering all of the three sensor placement scenarios stated above.
The covariance matrices for the pose and strain measurements are set as $\mbf{R}_p=\mathrm{diag}\left(10\sigma_t^2~10\sigma_t^2~10\sigma_t^2~10\sigma_a^2~10\sigma_a^2~10\sigma_a^2\right)$ and $\mbf{R}_s=\mathrm{diag}\left(10\sigma_\nu^2~10\sigma_\nu^2~10\sigma_\nu^2~10\sigma_\om^2~10\sigma_\om^2~10\sigma_\om^2\right)$ for pose and strain, respectively.
Similarly, the power spectral density matrix is set to  $\mbf{Q}_c=\mathrm{diag}\left(1~\mathrm{m}^2~1~\mathrm{m}^2~1~\mathrm{m}^2~100~\mathrm{rad}^2~100~\mathrm{rad}^2~100~\mathrm{rad}^2\right)$.
We note, that the comparably lower values in the translational part enforce a smoother rod with respect to shear and elongation, while the high values in the rotational part allow more deformation in bending and torsion.
These deformation properties and assumptions are very common in continuum robotics, as the manipulators often exhibit a low bending and torsional stiffness, while resisting shear and elongation forces.
An extreme case of this, where shear and elongation are completely neglected, is the so-called Kirchhoff rod.
Finally, we chose $K=29$ nodes for the state estimation algorithm, as a higher number of nodes does not seem to further increase the accuracy of the state estimation.
In addition, $M = 5$ interpolated states are used, in order to obtain a smooth representation of the robot shape.
The mean of the prior is initialized as a straight rod.

We note, that with the current choice of covariance matrices, we put relatively high trust into the measurements, while the prior cost are weighted relatively low in the optimization problem.
This way, the prior is mainly used for smoothing the solution of the regression problem, while sufficient sensor information is present to provide a good estimate about the resulting state and shape of the robot.

We further note, that tuning the hyperparameters is a non-trivial procedure as it depends on a number of different factors including the robot type, the type and number of sensors as well as the covariance and resulting confidence of these sensor readings.
While we chose to manually tune the hyperparameters in order to obtain desirable state estimates for the two-segment TDCR prototype in different sensor placement scenarios, future work should investigate more sophisticated optimization routines to identify these hyperparameters by using a larger set of calibration data.
For example, one could follow the approach proposed by \cite{wong_ral20b}, in which the parameters in a Gaussian variational inference setting are learned to obtain accurate state estimates of vehicle trajectories.

While tuning the hyperparameters for our TDCR design, we observed the resulting strain profiles of the simulated ground truth data.
Due to the nature and design of TDCR, the actuation loads that are acting on the robot are dominated by moments acting at discrete positions along the arclength at which the tendons of a segment terminate.
This results in a loading profile that is highly discontinuous, which also leads to a highly discontinuous strain profile with large jumps at the end of each respective TDCR segment.
An example of this is shown in Figure~\ref{fig:tdcr_strain_example}, which features a TDCR configuration that is mainly dominated by bending deformations around the local $y$-axis.
Both the $y$-component of the robot's orientation (represented as an angle-axis pair) and the $y$-component of the rotational strain are plotted along the arclength $s$.
It can be seen that the rotational strain is approximately constant for each individual bending segment.
It remains constant until the distal arclength $s = S$, where theoretically another discontinuity occurs, at which the strain jumps to zero.
When not locking the strain variables at $s = S$, the resulting estimated strain variable profile generally matches the simulated discontinuous one more closely, which results in slightly more accurate state estimates.
Thus, we did not lock the strain variables at $s = S$ during simulations and experiments in the following.
However, we note, that locking the strain at the tip of the rod is still the theoretically correct and widely used boundary condition for continuum robots.

\subsubsection{Results}

Detailed state estimation results for the configuration used to tune the hyperparameters are shown in Figure~\ref{fig:simulation_example}.
The estimated and ground truth robot shapes are plotted on the left, while the individual estimated and ground truth position and orientation components together with the 3$\sigma$ uncertainty envelopes are plotted on the right.
The three different sensor placement scenarios discussed above are considered, where the plots on the top consider two pose measurements at the end of each robot segment, the plots on the middle consider discrete strain measurements at each disk and the plots on the bottom consider strain measurements with a single pose measurement at the robot's tip.
It can be seen that in each of the three scenarios, the robot shape can accurately be estimated.
Larger deviations occur when only strain measurements are considered, which is to be expected as small errors and uncertainties are integrated along the arclength of the robot and can easily accumulate.
However, the ground truth remains close to the estimated shape and lies within the plotted uncertainty envelopes.
As expected, these envelopes are tighter when the estimation is more confident, which usually is the case, when an accurate measurement is present.
This behavior can most notably be observed in scenarios with pose measurements, in which both the envelopes for position and orientation tighten around the discrete measurements.
In the top plots, the envelops tighten at the end of each segment, while they only tighten at the tip of the robot in the bottom plots.
When considering only strain measurements, the uncertainty of the estimation increases along the robot's arclength, as expected.
Throughout all of the three different sensor placement scenarios convergence of the optimization routine occurred after three iterations, as indicated by the cost term plots at the bottom of Fig.~\ref{fig:simulation_example}.
Based on these results and the sparsity of the resulting linear system of equations, we anticipate, that real our proposed method can be utilized in real-time applications.

\begin{figure}[ht!]
	\centering
	\includegraphics[width=1\textwidth]{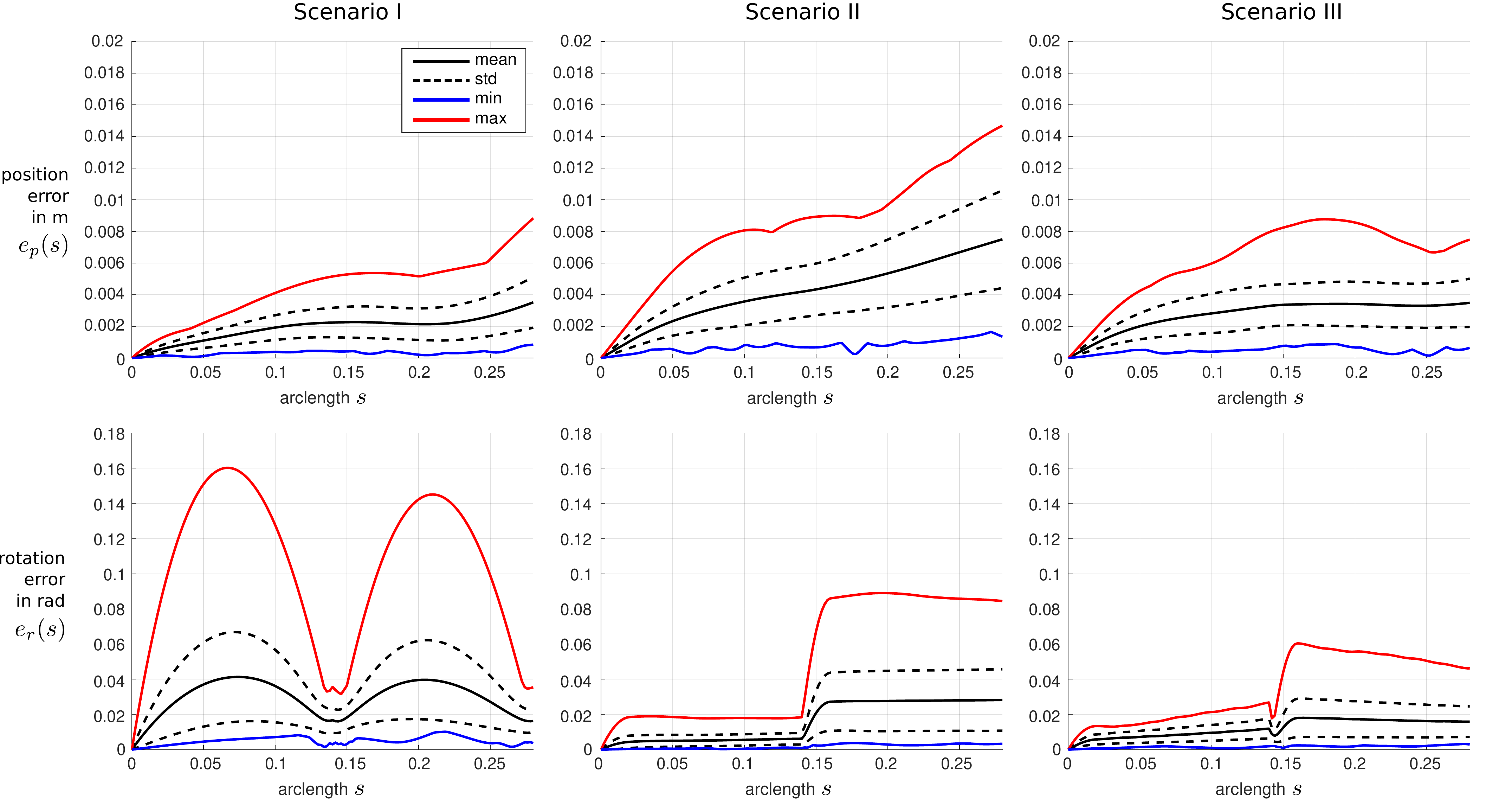}
	\caption{Position and orientation errors of the interpolated estimated state with respect to the simulated ground truth data along arclength $s$ considering all of the 100 simulated TDCR configuration from the sampled data set. The mean errors are shown in black, with the standard deviation plotted with dashed lines. The red and blue graphs show the maximum and minimum occurred errors at the corresponding arclength $s$. Three different sensor placement scenarios are considered: Pose measurements at the end of each robot segment (left), discrete strain measurements at each robot disk (middle), discrete strain measurements at each disk in combination with a single pose measurement at the robot's tip (right).}
	\label{fig:simulation_errors}
\end{figure}

For a more general evaluation, we run the state estimation approach for each of the 100 configurations included in our data set, while considering the previously described sensor placement scenarios.
For each run, we calculate both the position and orientation errors of the state estimation approach with respect to the simulated ground truth data along arclength $s$, considering $M = 5$ interpolated states between the $K$ nodes.
The position and orientation errors averaged over all configurations of our data set are plotted in Figure~\ref{fig:simulation_errors}.
Again we are considering the three sensor placement scenarios previously discussed: two discrete pose measurements, multiple discrete strain measurements and a single pose measurement at the tip ind combination with the strain measurements.

It can be seen that overall, relatively low errors between the shape estimation and the ground truth data occur.
Larger errors occur when only considering strain measurements, while pose measurements help to reduce the error significantly at the specific arclength positions.
In comparison to the results with strain measurements only, where the position error increases with respect to the robot arclength, pose measurements help to bound the error.
Overall, the average errors at the tip of the robot over all 100 configurations result in 3.5mm and 0.016$^\circ$ for the pose measurement scenario, 7.5mm and 0.028$^\circ$ for the strain measurement scenario, and 3.5mm and 0.016$^\circ$ for the pose and strain measurement scenario.
We do note that there will always be some error between the estimation and ground truth, since noisy sensor data is used in the estimation algorithm.

\subsection{Experiments}

The proposed algorithm is further evaluated using real robot experiments.
In order to do so, a two-segment TDCR prototype is built using the parameters stated in Table~\ref{tab:tdcr_params}.
The prototype consists of a super-elastic Nitinol backbone and spacer disks, that are used to route four tendons per segment along the robot's backbone.
The spacer disks are fabricated using a stereolithography resin 3D printer (Form 3, Formlabs Inc., USA).
The robot can be actuated by pulling different combinations of those tendons.
The prototype is shown in Figure~\ref{fig:experimental_setup} on the left.

\subsubsection{Sensor Setup}

For the experimental evaluation, we consider the sensor placement scenario with two discrete pose measurements, one at the end of each robot segment.
In order to obtain these pose measurements, an electromagnetic tracking system (Aurora v3, Northern Digital Inc., Canada) is used.
Two custom 6 degree-of-freedom tracking coils are rigidly attached to the spacer disks at the end of the corresponding robot segment.
A third tracking sensor is attached to the base of the robot to serve as 6-degree-of-freedom reference frame.
Using  its  measured  position  and  orientation  data  and  the known  transformation  between  the  sensor  mounting  point and  the  base  of  the  robot,  the  transformation  between  the electromagnetic tracking system and the robot’s base frame can  be  calculated.
The used tracking sensors have a stated root mean square error of 0.8 mm and 0.7$^{\circ}$ by the manufacturer.

In addition to the sensor readings from the electromagnetic tracking system, the shape of the robot is reconstructed using a highly accurate external laser scan.
This data serves as ground truth in order to assess the accuracy of the proposed state estimation algorithm.
The shape is represented as a frame of each of the robot's disks.
The robot is scanned using a coordinate measurement arm with attached laser probe (FARO Edge with FARO Laser Line Probe HD, FARO Technologies Inc., USA).
The laser probe allows to obtain accurate point cloud readings of the robot's shape.
Each printed disk holds three spheres with diameters of 5mm, that can be easily identified and extracted from the corresponding point cloud.
Using the position of these three spheres and the disk's known geometry, the frame holding the spatial position and orientation can be reconstructed for each.
On top of that, the robot's base also contains three spheres, to identify its frame.
Thus, the transformation between the measurement arm system and the robot's base frame can be computed.

\begin{figure}[t!]
	\centering
	\includegraphics[width=1\textwidth]{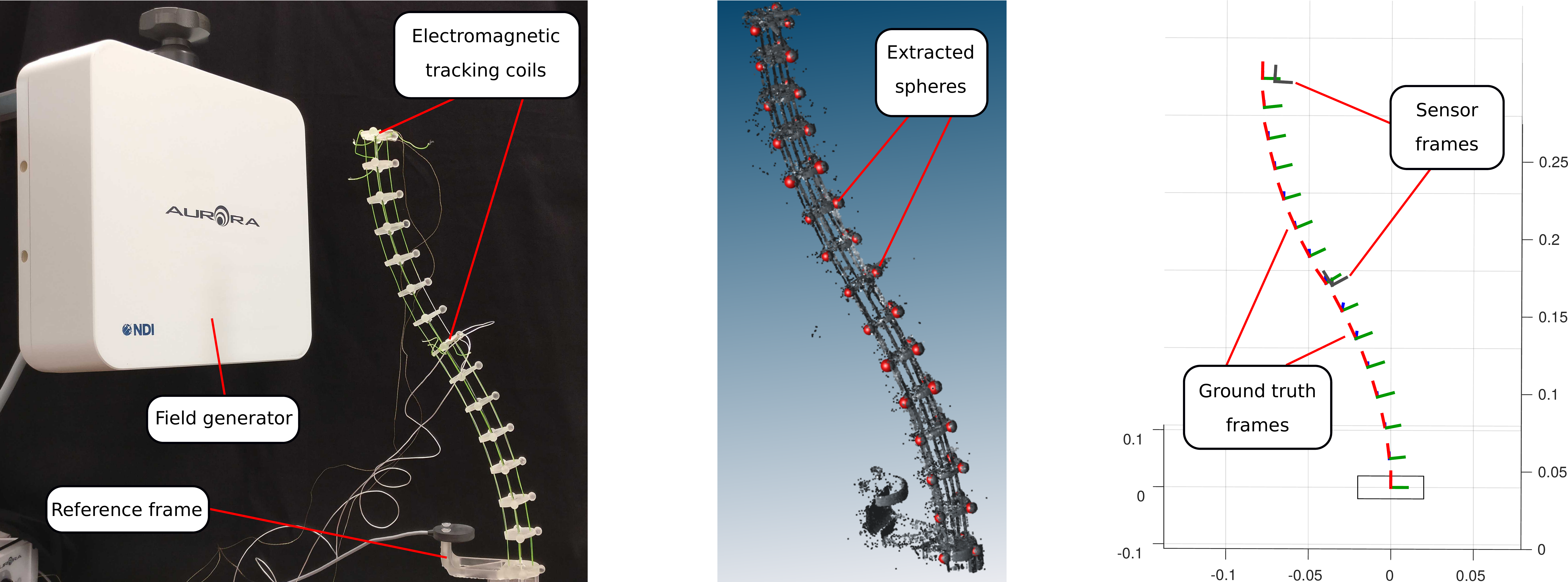}
	\caption{Experimental setup including the TDCR prototype and the electromagnetic tracking system (left); Laser scan of the TDCR prototype with spheres extracted from the point cloud to reconstruct the frame of each disk (middle); extracted colored ground truth frames of each disk from the laser scan together with two gray disk frames measured by the electromagnetic tracking system (right).}
	\label{fig:experimental_setup}
\end{figure}

The experimental setup as well as the extraction of ground truth and sensor frame is shown in Figure~\ref{fig:experimental_setup}.
Finally, both the electromagnetic sensor readings and the point cloud extractions can both be represented with respect to the robot's base frame, allowing it to use them in the state estimation algorithm and compare them with respect to each other.
However, the calibration between the two distinct measurement systems might be slightly inaccurate for multiple reasons.
First, errors might occur in the calculated transformation between the electromagnetic sensing system and the robot's base frame arising from uncertainties in sensor placement and inaccuracies in the 3d printed base frame.
Second, the accuracy of the transformation calculated for the ground truth data obtained from laser scans highly depends on the accuracy of the sphere extraction.
In order to overcome these inaccuracies and to better align the two measurement systems, an optimization based calibration routine is employed in the later sections.

\begin{figure}[ht!]
	\centering
	\includegraphics[width=0.9\textwidth]{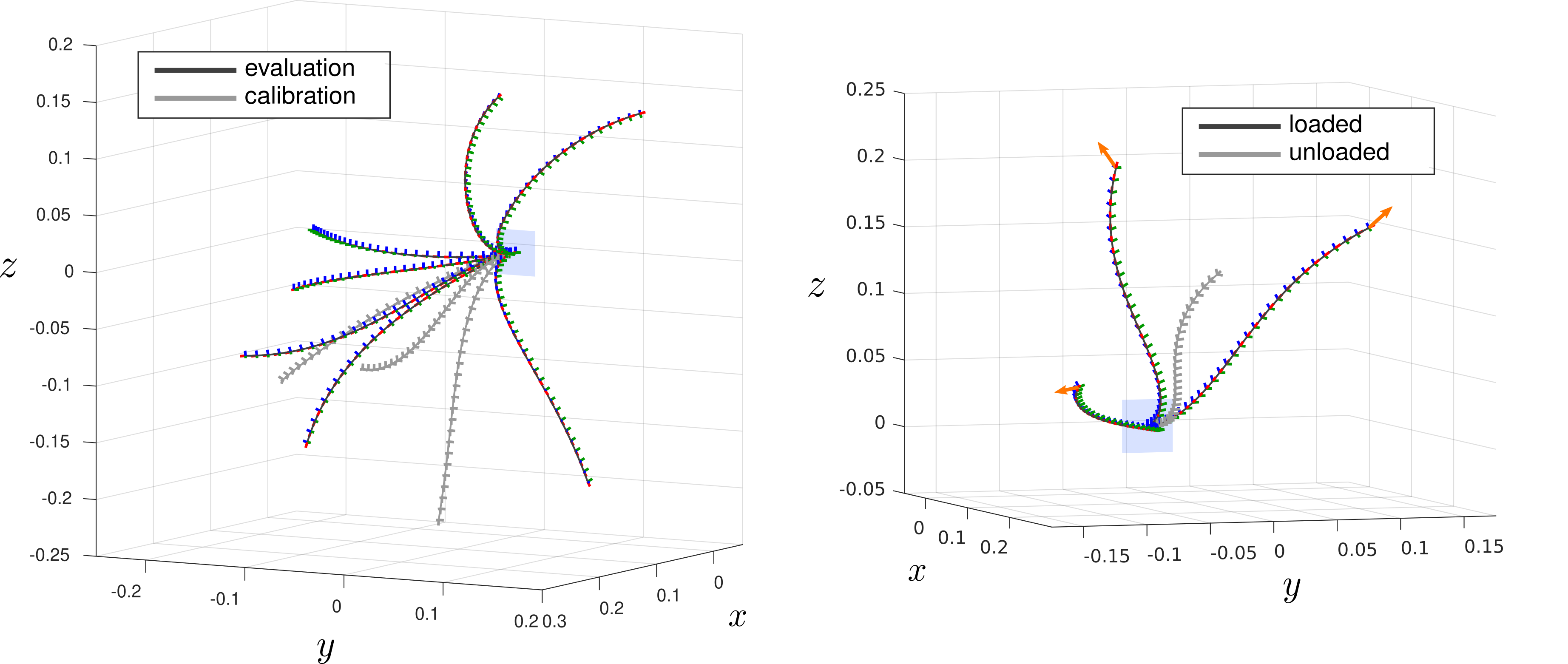}
	\caption{Left: Shapes of nine TDCR configurations considered for the experimental validation of the state estimation approach. Three configurations (gray) were used to calibrate the sensing system to the ground truth data obtained from the laser scan, while the remaining six configurations (colored) are used to assess the performance of the shape estimation algorithm. Right: Shapes of three loaded TDCR configurations (colored), resulting from applying a tip load to a TDCR configuration in different directions. The initial unloaded configuration is plotted in gray and the directions of the applied loads are shown in orange.}
	\label{fig:robot_configs}
\end{figure}

\subsubsection{Data Collection}

Nine different TDCR configurations are used for the experimental validation.
These configurations are obtained by actuating different combinations of tendons of the continuum robot, leading to different shapes covering different areas of the robot's workspace.
For each of these configurations, the two pose measurements from the electromagnetic sensors at the end of each segments are recorded.
In addition, the ground truth disk frames are extracted using a laser scan.
The resulting robot shapes for each configuration are plotted in Figure~\ref{fig:robot_configs} (left).
In the following, three out of these nine configurations are used to further calibrate the transformation between the individual sensing systems and the robot base frame.
These three configurations are shown in gray in the plot.
The remaining six configurations are then used to evaluate the proposed state estimation approach.

For further evaluation of our proposed method, we are also considering three TDCR configurations, in which an additional force of approximately 0.2 N is applied to the robots tip.
The resulting robot shapes for these configurations are plotted in Figure~\ref{fig:robot_configs} (right).
The initial, unloaded TDCR configuration is plotted in gray, while the shapes of the three loaded configurations are colored.
The directions of the applied tip forces are plotted in orange.

We note, that the number of evaluation configurations is relatively low, mainly due to the time consuming manual process of extracting the ground truth shape and frames from the laser scan.

\subsubsection{Calibration}

As stated above, we aim to further calibrate the alignment of the two individual measurement systems used throughout the experimental evaluation.
We achieve this by employing a numerical optimization scheme using three of the nine recorded unloaded TDCR configurations.
In particular, we aim to minimize the offsets between the ground truth frames and the frames obtained from the electromagnetic sensors for the two disks at the end of the robot's segments.
Thus, we have two data points per configuration, totaling in six data points for this optimization problem.
We want to find the transformation matrix between the measurement arm laser probe base frame and the base frame of the electromagnetic tracking system that minimizes the average position error between the two measurement systems for the six data points.
We implement a simple numerical optimization routine in \textsf{MATLAB} using \texttt{fsolve}.
After optimization the remaining average position error for the six data points between the electromagnetic sensor readings and the laser scan reconstructions results in 1.2 mm.
The three TDCR configuration used for this calibration are discarded from the data set and are not considered when evaluating the state estimation approach in the following.

\begin{figure}[ht!]
	\centering
	\includegraphics[width=1\textwidth]{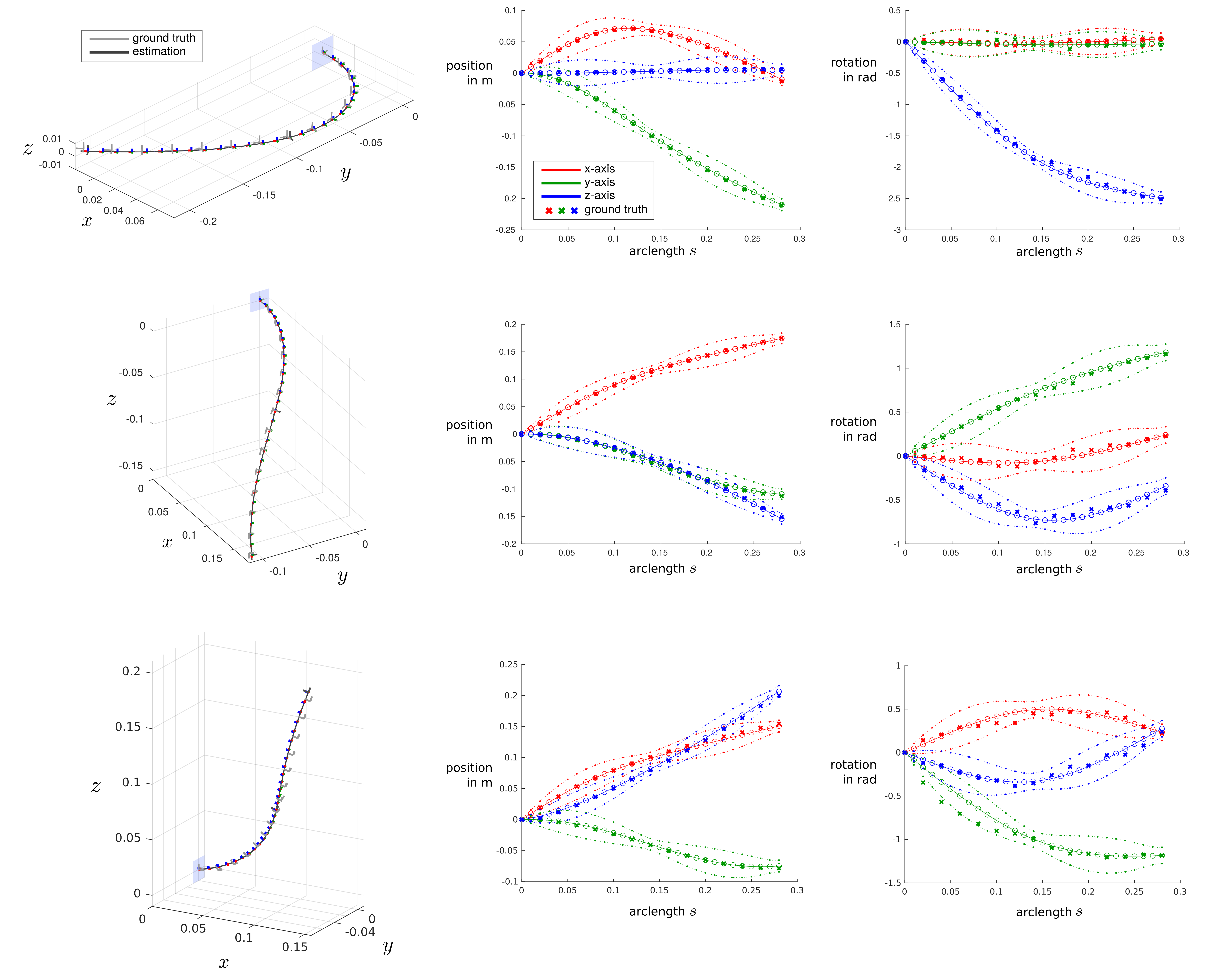}
	\caption{Example results from the proposed state estimation algorithm using three of the evaluated configuration of the TDCR prototype - the top two configurations are unloaded, while the bottom configuration has an additional force applied to its tip. The resulting continuous estimated shapes are plotted using colored frames on the left with ground truth frames plotted in gray. The plots on the right display the the $x$- (red), $y$- (green) and $z$-components (blue) of the robot's translation and orientation (expressed as angle-axis pair) along its arclength $s$. The plots show the mean along with the 3$\sigma$ uncertainty envelopes and the ground truth data from the laser scans (cross marks).}
	\label{fig:experiment_example}
\end{figure}

\subsubsection{Results}

The state estimation algorithm was run on the remaining six configurations of the TDCR prototype.
The hyperparameters remained the same as tuned in Section~\ref{sec:hyperparameter} to remain comparable to the results obtained in simulation.
The only exception is the number of interpolated states, which was changed to $M=2$.
We run the state estimation approach for each of the six TDCR configurations, while considering the previously described sensor placement scenario, in which the pose of the end and middle disk of the robot are measured using electromagnetic tracking sensors.

\begin{figure}[ht!]
	\centering
	\includegraphics[width=0.8\textwidth]{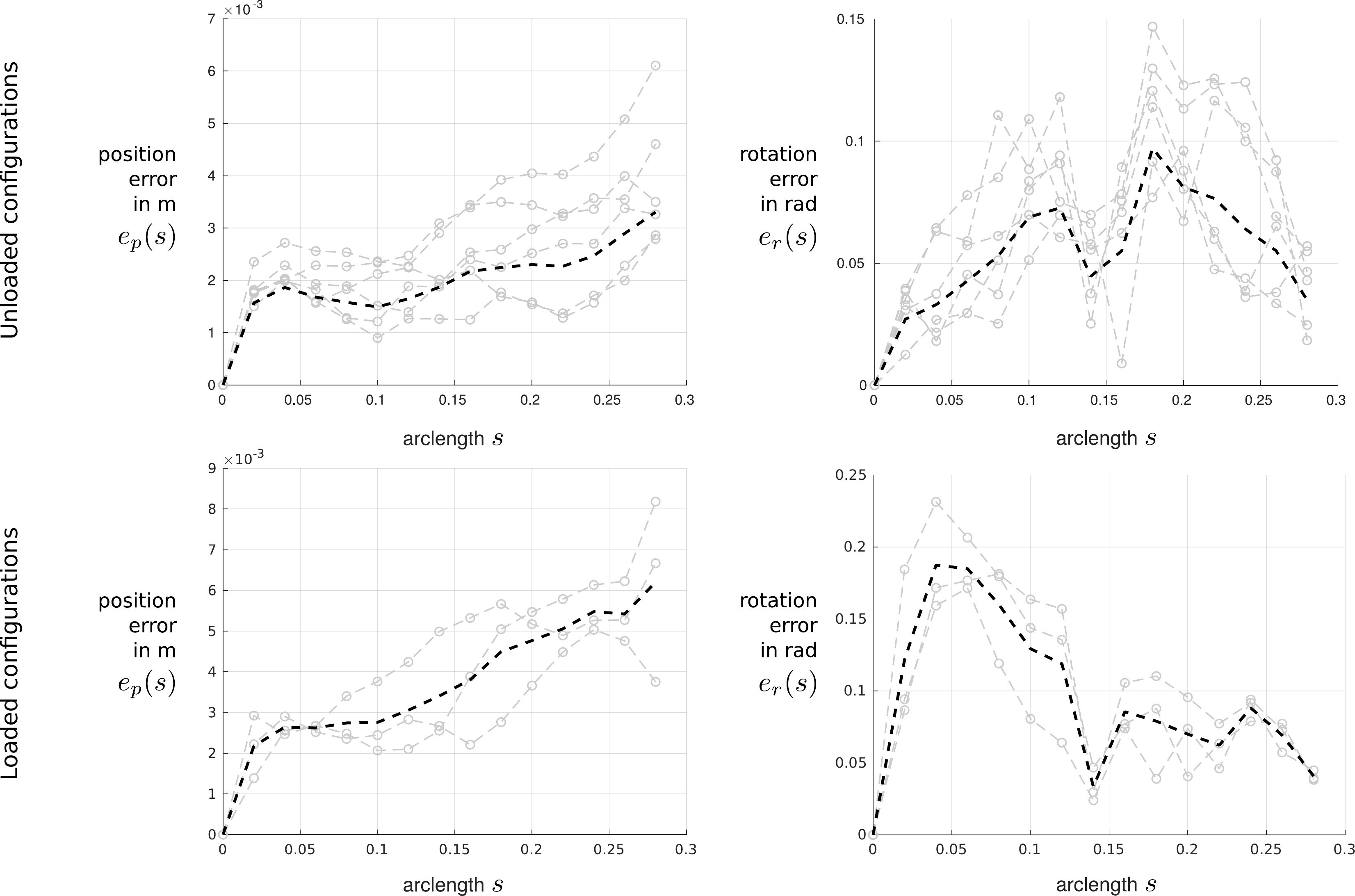}
	\caption{Position and orientation errors of the estimated state with respect to the ground truth data obtained from the laser scan along arclength $s$. Top: Six unloaded TDCR configurations are plotted in gray, while the mean errors across all six configurations are shown in black. Bottom: Three TDCR configurations with an additional tip load are plotted in gray, while the mean errors across all six configurations are shown in black.}
	\label{fig:arclength_errors_experiments}
\end{figure}

Detailed state estimation results for three example configurations are shown in Figure~\ref{fig:experiment_example}, one of which has a force applied to its tip (bottom).
The estimated and ground truth robot shapes are plotted using on the left, while the individual estimated and ground truth position and orientation components together with the 3$\sigma$ uncertainty envelopes are plotted on the right.
Agreeing with the results from simulations, the robot shape can accurately be estimated.
The ground truth measurements remain close to the estimated shape and lies within the plotted uncertainty envelopes.
Again, these envelopes are tighter when the estimation is more confident, which usually is the case, when an accurate measurement is present.

The position and orientation errors averaged over all of the six unloaded and three loaded TDCR configurations are shown in Figure~\ref{fig:arclength_errors_experiments}.
It can be seen that overall, relatively low errors between the shape estimation and the ground truth data occur.
Specifically, average end-effector errors of 3.3mm and 0.035$^\circ$ are achieved for the unloaded configurations.
For the loaded configurations, the average end-effector errors result in 6.2mm and 0.041$^\circ$.
These errors and their overall trend agrees very well with the results seen in simulation.
However, orientation of the estimated shape appears to be less accurate than the position when compared to the ground truth, as larger deviations are present.
We note, that the orientation of the reconstructed ground truth disk frames from the laser scan might be slightly inaccurate due to slight misalignments between the disks and the backbone during manufacturing as well as inaccuracies during sphere extraction.
Thus, the errors might not be solely attributed to the state estimation algorithm.

\section{Discussion}

The results presented in the previous section indicate that accurate shape estimates based on limited sensor information can be achieved by the proposed state estimation approach, resulting in average position and orientation errors at the end-effector as low as 3.5mm and 0.016$^\circ$ in simulation or 3.3mm and 0.035$^\circ$ for unloaded configurations and 6.2mm and 0.041$^\circ$ for loaded ones during experiments, when using discrete pose measurements.
The evaluation carried out in this work is limited to a multisegment TDCR.  However, the general formulation of the utilized Cosserat rod model makes the method applicable to a large variety of continuum robots.
The main advantages of the proposed approach include the continuous estimation of the robot's state, allowing for easy interpolation between nodes (mean and covariance), as well as the sparse formulation of the GP regression, requiring little computational effort.

We note, that in the current formulation of the prior will always lead to continuous profiles for the continuum robot's estimated state, including both pose and internal strains along the arclength $s$.
However, the presence of a discrete point load acting on the robot structure at a particular location along $s$ can in fact lead to a discontinuous strain profile.
One example of this can be seen in the multisegment TDCR considered in this work.
The tendons routed along the robot's structure apply discrete bending moments at their termination location, leading to a sudden and discontinuous change of the internal bending strain.
In such cases, the state estimation approach will fit a smooth and continuous strain profile, which will differ from the ground truth and can lead to inaccuracies in the estimation.
An example of this can be seen in the simulation results using strain measurements presented in Fig.~\ref{fig:simulation_errors}.
In this case, the orientation error increases significantly at the end of the first segment, which is located at half of the robot's total length.

\begin{figure}[ht!]
	\centering
	\includegraphics[width=0.8\textwidth]{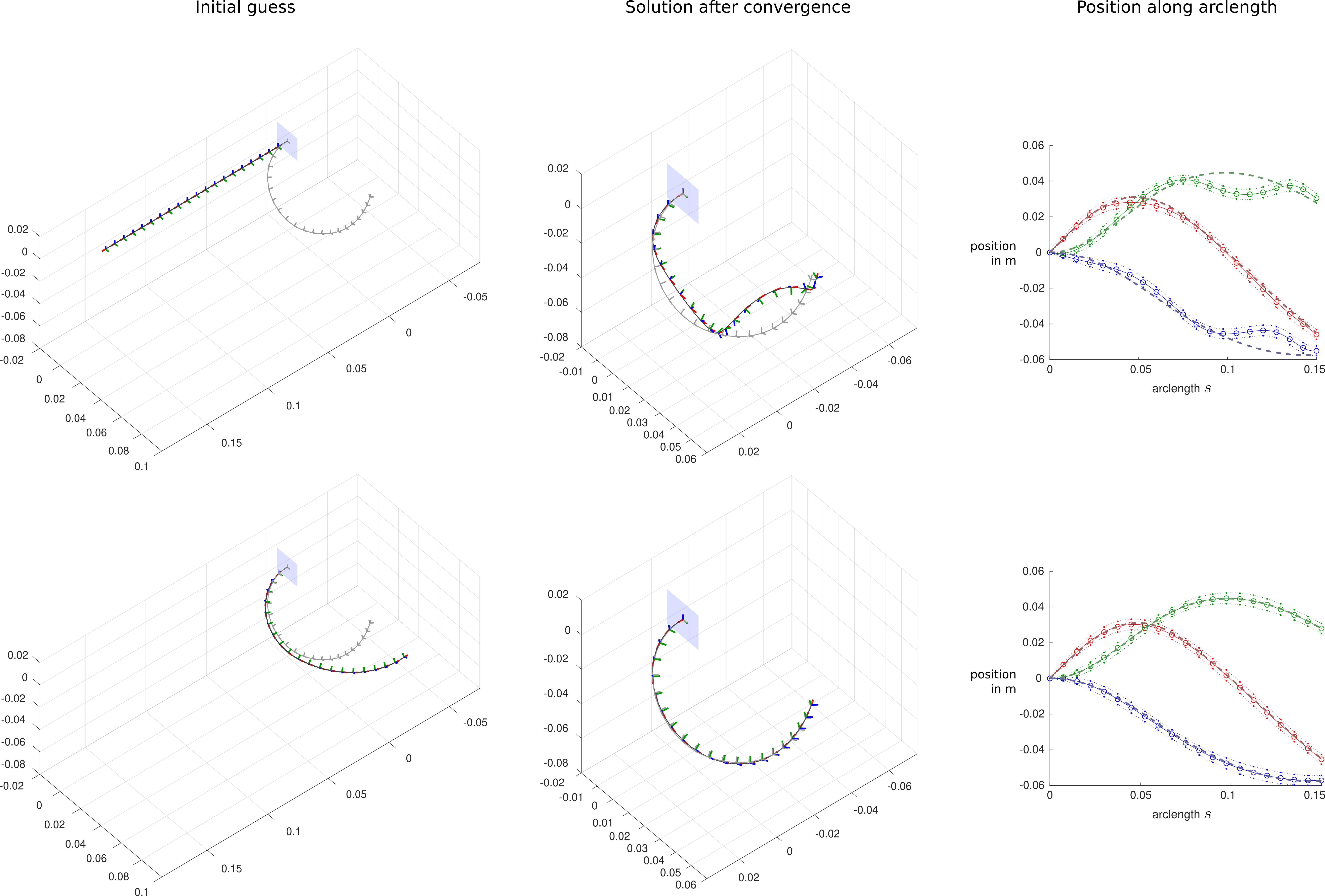}
	\caption{Effects of the initial guess on the convergence of the optimization problem: An initial guess (colored robot shape) that is not close to the ground truth data (gray robot shape) or the associated sensor readings, can result in converging to a local minimum, leading to shape estimation solutions that do not reflect the ground truth (top); A possible solution to that problem is to choose an initial guess closer to the ground truth, e.g. obtained from a kinematic model of the corresponding robot, leading to proper convergence (bottom).}
	\label{fig:initial_guess}
\end{figure}

Further, we would like to point out, that the approach presented in this work is a local optimization method and therefore heavily depends on the chosen initial guess.
A poorly chosen initial guess, which is far away from the robot's ground truth and the corresponding sensor measurements, can lead to convergence to a local minima, that does not necessarily align with the expected robot shape.
This problem could be overcome by using a more informed initial guess than the initially straight shape that is used throughout this paper.
This initial guess could for instance be obtained from a kinetostatic model, relating known robot actuation inputs to a resulting robot shape.
An example of this is shown in Fig.~\ref{fig:initial_guess} for a two segment TDCR.
In this example, the ground truth shape of the robot results from tendon actuation with an additional external force at the robot's tip and is highly curved.
Two pose measurements, one at the end of each robot segment, are used to estimate the robot's shape with the approach proposed in this paper.
Using a straight shape as an initial guess (top), results in convergence to a local minima, that aligns with the sensor readings but not with the expected ground truth shape.
Using the shape prediction of a Cosserat rod based model for TDCR as an initial guess instead (bottom), assuming that tendon actuation values are known, convergence to the expected ground truth shape can be achieved.

Finally, we would like to emphasize and acknowledge that our current prior is based on simplified assumptions, as it does not incorporate any knowledge about external or actuation forces.
This knowledge could be incorporated in our prior by assuming a Gaussian-process $\mbf{w}(s)$ with a non-zero mean function, which in return would lead to improved state estimates, especially when less accurate sensor readings are present.
However, in the scenarios considered throughout this paper, sufficient sensor information is available, making the simplified prior adequate to obtain high quality state estimates as evident from the presented results.

\section{Conclusion}

This paper describes a state estimation approach for a continuum robot utilizing sparse Gaussian process regression.
The proposed framework makes use of a simplified Cosserat rod formulation, making it applicable to a large variety of continuous manipulator, without requiring additional robot-dependent knowledge and information.
By considering limited discrete sensor information, the approach allows for a continuous estimation of the robot's full state, including backbone pose and internal strains as well as their respective uncertainty envelopes.
Evaluation with simulations and real robot experiments show that accurate and continuous estimates of a continuum robot's shape can be achieved, resulting in average end-effector errors between the estimated and ground truth shape of 3.5mm and 0.016$^\circ$ in simulation or 3.3mm and 0.035$^\circ$ during experiments for unloaded configurations and 6.2mm and 0.041$^\circ$ for loaded ones, when using discrete pose measurements.

Future work could focus on extending the proposed approach to also include and estimate the second derivative of the backbone's pose and position, which would enable estimation of the forces and moments acting on the robot structure.  Furthermore, although we have demonstrated the method on a single continuum robot, we believe it will be possible extend the method to parallel continuum robots quite easily where there are loops in the topology; by viewing this as a batch state estimation problem, we can again borrow ideas from mobile robotics to perform inference over topology graphs with loops.

\section*{Acknowledgements}

The authors are indebted to Gabriele D'Eleuterio, also at the University of Toronto, for helpful discussions around the similarity between the rigid-body and Cosserat equations expressed using $SE(3)$ tools.  We also appreciate the help of David Yoon at the University of Toronto, who independently verified our numerical implementation of the state interpolation equations.  This work was supported by the Natural Sciences and Engineering Research Council (NSERC) of Canada.

\section*{Declaration of Conflicting Interests}

The Authors declare that there is no conflict of interest.

\section*{Funding}

This research received no specific grant from any funding agency in the public, commercial, or not-for-profit sectors.

\bibliographystyle{asrl}
\bibliography{refs,refs2,book,pubs,refs_continuum}

\end{document}

%% file: notation_header.tex
\newcommand{\bbm}{\begin{bmatrix}}
\newcommand{\ebm}{\end{bmatrix}}
\newcommand{\mbf}{\mathbf}
\newcommand{\mbs}[1]{{\boldsymbol{#1}}}

\def\om{\omega}

\newcommand{\beq}{\begin{equation}}
\newcommand{\eeq}{\end{equation}}
\newcommand{\bdis}{\begin{displaymath}}
\newcommand{\edis}{\end{displaymath}}
\newcommand{\beqn}[1]{\begin{subequations}\label{eq:#1}\begin{eqnarray}}
\newcommand{\eeqn}{\end{eqnarray}\end{subequations}}
\newcommand{\est}[1]{\hat{#1}}
\newcommand{\pri}[1]{\check{#1}}
\newcommand{\wdg}{\wedge}

\newcommand{\Wdg}{\curlywedge}

\newcommand{\Tbig}{\,\mbox{\boldmath ${\cal T}$\unboldmath}}

\newcommand{\Tsmall}{\mbf{T}}

\newcommand{\Jbig}{\mbs{\mathcal{J}}}